\documentclass[10pt,journal,compsoc]{IEEEtran}

\ifCLASSOPTIONcompsoc
 \usepackage{cite}
\else
  \usepackage{cite}
\fi

\ifCLASSINFOpdf
\else
\fi
\usepackage{amsmath,amssymb}
\usepackage{multirow}
\usepackage{booktabs}
\usepackage{algorithmic}
\usepackage{algorithm}
\usepackage{subcaption}
\usepackage{graphicx}
\usepackage{xcolor}
\usepackage{verbatim}

\newtheorem{theorem}{Theorem}[section]

\newtheorem{proposition}[theorem]{Proposition}
\newtheorem{lemma}[theorem]{Lemma}
\newtheorem{remark}[theorem]{Remark}
\newtheorem{example}[theorem]{Example}
\newtheorem{corollary}[theorem]{Corollary}

\newenvironment{Theorem}{\begin{theorem}\sl}{\end{theorem}}

\newenvironment{Proof}{\textbf{Proof:}}{\hfill$\Box$\\}

\begin{document}
%
\title{GO-LDA: Generalised Optimal Linear Discriminant Analysis}
%
%

%
%
 
\author{Jiahui Liu,
        Xiaohao Cai,
        and~Mahesan Niranjan 
\thanks{
The authors are with the School of Electronics and Computer Science, University of Southampton, Southampton SO17 1BJ, UK. \\
E-mail: jl4f19@soton.ac.uk; x.cai@soton.ac.uk; mn@ecs.soton.ac.uk}
}

\IEEEtitleabstractindextext{%
\begin{abstract}
Linear discriminant analysis (LDA) has been a useful tool in pattern recognition and data analysis research and practice. While linearity of class boundaries cannot always be expected, nonlinear projections through pre-trained deep neural networks have served to map complex data onto feature spaces in which linear discrimination has served well. Unlike principal component analysis which is variance preserving, LDA maximises the separation between classes, simultaneously minimising the projected scatter of each class on a subspace. The solution to binary LDA is obtained by eigenvalue analysis of within-class and between-class scatter matrices. It is well known that the multiclass LDA is solved by an extension to the binary LDA, a generalised eigenvalue problem, from which the largest subspace that can be extracted is of dimension one lower than the number of classes in the given problem.
In this paper, we show that, apart from the first of the discriminant directions, the generalised eigenanalysis solution to multiclass LDA does neither yield orthogonal discriminant directions nor maximise discrimination of projected data along them. Surprisingly, to the best of our knowledge, this has not been noted in decades of literature on LDA. 
To overcome this drawback,
we present a derivation with a strict theoretical support for sequentially obtaining discriminant directions that are orthogonal to previously computed ones and maximise in each step the Fisher criterion. We show distributions of projections along these axes and demonstrate that discrimination of data projected onto these discriminant directions has optimal separation, which is much higher than those from the generalised eigenvectors of the multiclass LDA. 
Using a wide range of benchmark tasks, we present a comprehensive empirical demonstration that on a number of pattern recognition and classification problems, the optimal discriminant subspaces obtained by the proposed method, referred to as GO-LDA (Generalised Optimal LDA), can offer superior accuracy.
\end{abstract}

}

\maketitle

\IEEEdisplaynontitleabstractindextext

%
\IEEEpeerreviewmaketitle

\section{Introduction}\label{sec:introduction}

Linear discriminant analysis (LDA) has been a widely used technique for pattern recognition over several decades starting from the seminal work of Fisher \cite{fisher1936use}. While recent large-scale problems, such as computer vision, are solved by deep neural networks, where do exist several problems of industrial and societal importance posed on relatively small data sets for which LDA is still the effective tool of choice. LDA, in a vast majority of problems such as screening for a medical condition solving binary classification problems, seeks to find a direction in the space of features such that Fisher criterion of separation between projected means is maximised and simultaneously the within-class scatter of projections is kept low. Foley and Sammon in \cite{foley1975optimal} extended LDA in the binary regime by constructing a discriminant subspace following the derivation of discriminant directions. It begins with the first discriminant direction given by maximising the Fisher criterion and sequentially constructing discriminant directions which also maximise discrimination information but are constructed to be orthogonal to those previously derived.

\begin{figure*}[h]
\centering
\begin{minipage}[c]{0.6\textwidth}
\includegraphics[width=4.0in]{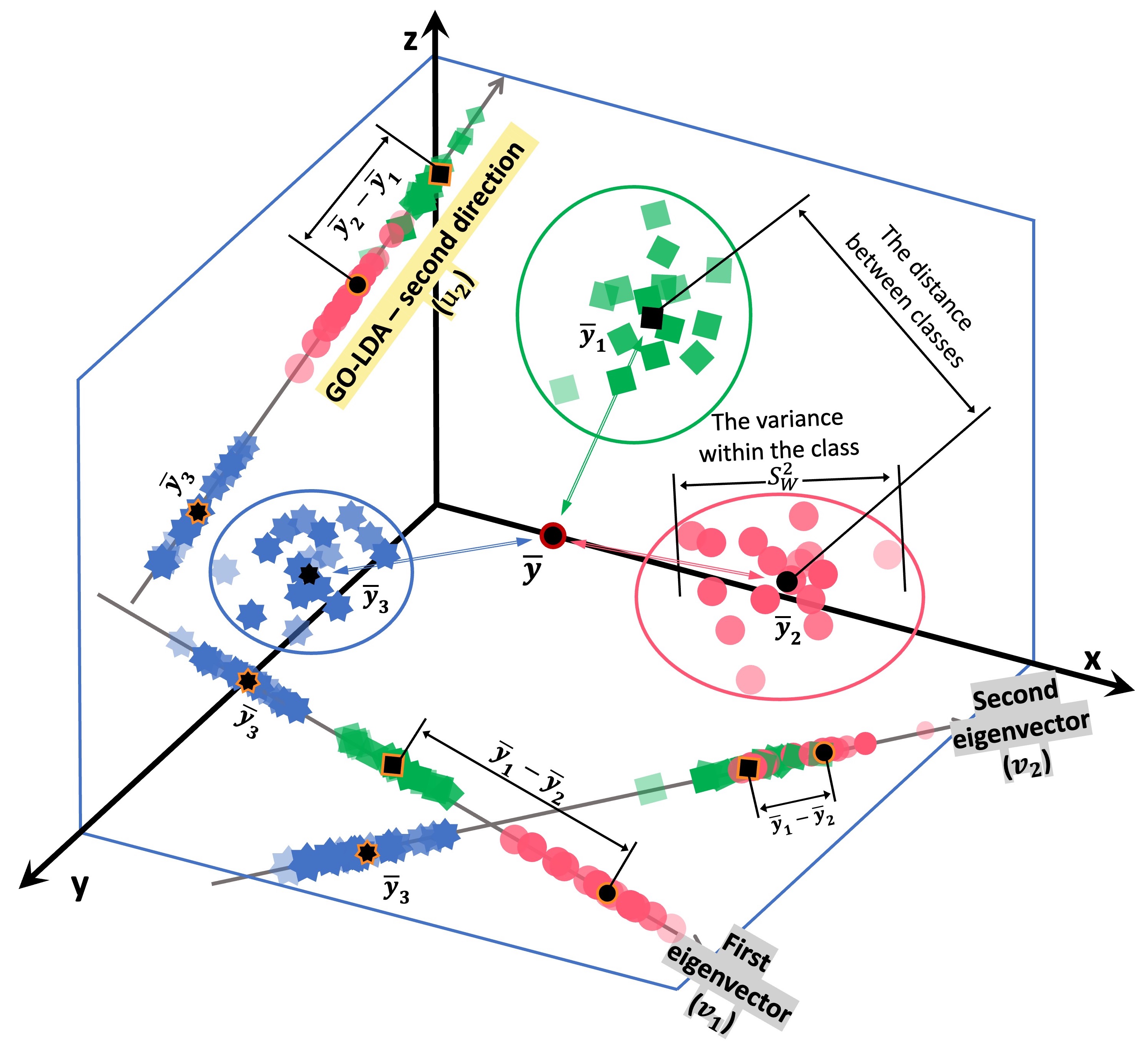}
\end{minipage}
\hspace{-0.13in}
\begin{minipage}[c]{0.30\textwidth}
\caption{Schematic diagram for a three-class problem showing the difference between the classic-LDA (generalized eigenanalysis solution) and the proposed GO-LDA (sequential computation of discrimination maximising orthogonal projections). Along the direction of the first eigenvector $\boldsymbol{v}_{1}$, the three classes are separated well, but the second direction (i.e., $\boldsymbol{v}_{2}$) of the classic-LDA is not orthogonal to the first and fails to separate two of the three classes, whereas a better solution exists, i.e., the direction (i.e., $\boldsymbol{u}_{2}$) of our GO-LDA. For ease of reference, the symbols in the diagram for the class means, i.e., 
$\{\bar{\boldsymbol{y}}_j\}_{j=1}^3$, are reused for the mean of the projected samples in each class.
}
\label{fig:go-lda-diagram}
\end{minipage}
\end{figure*}

Fisher formulation has a multiclass extension for which the solution is arrived at by solving a generalised eigenvalue problem involving the sum of within-class scatter matrices and between-class matrices obtained on the sum of outer products of vectors linking the pairwise means \cite{rao1948utilization}. A consequence of this is that the dimension of the obtained discriminant subspace is limited to one less than the number of classes of the given problem \cite{bishop2006pattern}.

An aspect of the multiclass LDA, formed by generalised eigenvectors, is that apart from the first of the directions, the others neither maximise discrimination nor are they orthogonal to each other. To the best of our knowledge, this particular fact/drawback has been overlooked in the LDA literature for decades. In this paper, we show this to be the case and, building on Foley and Sammon's work \cite{foley1975optimal}  on sequential construction, derive orthogonal discriminant directions for multiclass LDA. This result is an optimal subspace on which to project data while maintaining their separation. We present the necessary algebraic derivations, illustrative examples showing the distributions of projected data and a comprehensive set of experiments to demonstrate how successive directions computed carry discriminant information and that combination offers subspace in which simple classifiers (e.g., linear, quadratic and $k$-nearest neighbour) would be applied. Another important result of the work is that in the case of multiclass problems, we are no longer limited to the discriminant subspace being of a smaller (one fewer) size than the number of classes of the given problem.
We refer to our work as GO-LDA --Generalised Optimal LDA; see Fig. \ref{fig:go-lda-diagram} for a brief illustration of the difference between the multiclass LDA (i.e., classic-LDA) and GO-LDA on a three-class problem.

Our derivation, supported by empirical work, shows that a sequential construction of mutually orthogonal discriminant directions finds a linear subspace in which greater accuracy of classification may often be obtained for multiclass problems. Surprisingly, this construction goes beyond the limit set by the number of classes of the given problem owing to the rank deficiency of the between-class scatter matrix encountered in the classic-LDA formulation. We note, however, consistent with the {\em no free lunch theorem}, that accuracy gains cannot always be expected for two reasons. Some problems may be sufficiently easy that a small number of discriminant directions (as few as one) might carry all useful information. At the other extreme, some problems might require non-linear classification boundaries that cannot be modelled by linear projections. Nevertheless, it is an intriguing finding and of paramount importance that for several multiclass problems there still remains unextracted information even with linear projections. Using a wide range of benchmark tasks, we show that GO-LDA outperforms the classic-LDA and other related methods by a large margin. It can also be exploited with deep learning techniques and can achieve state-of-the-art performance.

The remainder of this paper is organized as follows. Section \ref{notation} defines the necessary  notation. A brief survey of discriminant analysis is given in Section \ref{sec:related_work}.
Section \ref{sec:methodology} recalls the mathematical aspect regarding LDA. In Section \ref{sec:proposed_method}, we present the proposed methodology GO-LDA, including mathematical derivations, computational complexity analysis and some illustrations of the discriminant ability and projections of data onto the computed discriminant directions. Extensive experimental results and comparisons validating the superior performance of GO-LDA are given in Section \ref{sec:experiments}. We finally conclude with a discussion in Section \ref{sec:conclusion}. Further related derivation and illustrations are given in Appendix.

\section{Notations}
\label{notation}
Given $N$ samples $\boldsymbol{y}_i = (y_{i1}, y_{i2}, \cdots, y_{iM})^\top \in \mathbb{R}^M, 1\le i \le N$, we form a data matrix $\boldsymbol{Y} = (\boldsymbol{y}_1, \boldsymbol{y}_2, \cdots, \boldsymbol{y}_N)^\top \in \mathbb{R}^{N\times M}$, where $M$ is the number of features of every sample. Suppose that these $N$ samples belong to $C$ different classes, namely $\boldsymbol{\Lambda}_j$, and their cardinality $|\boldsymbol{\Lambda}_j| = N_j$, $1\le j \le C$. Let $\boldsymbol{y}^j_k$ represent the $k$-th sample in class $\boldsymbol{\Lambda}_j$. Clearly, $N = \sum_{j=1}^C N_j$, $\boldsymbol{\Lambda}_j = \{\boldsymbol{y}^j_k\}_{k=1}^{N_j}$ and $\{\boldsymbol{y}_i\}_{i=1}^N = \bigcup_{j=1}^C \{\boldsymbol{y}^j_k\}_{k=1}^{N_j}$. Let $\bar{\boldsymbol{y}}$ and $\bar{\boldsymbol{y}}_j$ respectively be the mean of the whole samples and the samples in class $j$, i.e., $\bar{\boldsymbol{y}} = \frac{1}{N} \sum_{i=1}^N \boldsymbol{y}_i$, $\bar{\boldsymbol{y}}_j = \frac{1}{N_j} \sum_{\boldsymbol{y} \in \boldsymbol{\Lambda}_j} \boldsymbol{y}$, $1\le j \le C$.

Denote $\boldsymbol{S}_{\rm B}$ and $\boldsymbol{S}_{\rm W}$, respectively, as the inter- and intra-class (also known as between- and within-class) scatters, i.e.,
\begin{equation}
\boldsymbol{S}_{\rm B}  =\sum_{j=1}^C(\bar{\boldsymbol{y}}_j - \bar{\boldsymbol{y}})(\bar{\boldsymbol{y}}_j - \bar{\boldsymbol{y}})^{\top}, \quad
\boldsymbol{S}_{\rm W} =\sum_{j=1}^C  \boldsymbol{S}^j_{\rm W},
\label{SB_SW}
\end{equation}
where $\boldsymbol{S}^j_{\rm W}$ represents the intra-class scatter for class $j$, i.e.,
\begin{equation}
\boldsymbol{S}^j_{\rm W}= \sum_{k=1}^{N_{j}}(\boldsymbol{y}_k^j-\bar{\boldsymbol{y}}_j) (\boldsymbol{y}_k^j-\bar{\boldsymbol{y}}_j)^{\top}, \ \ 1\le j \le C.
\end{equation}
Specifically, for the case of $C=2$, we also name $\tilde{\boldsymbol{S}}_{\rm B}$ as the inter-class scatter, i.e.,
\begin{equation} \label{eqn:bi-intra-inter}
\tilde{\boldsymbol{S}}_{\rm B} = \boldsymbol{s}_{\rm b} \boldsymbol{s}_{\rm b}^\top,
\end{equation}
where $\boldsymbol{s}_{\rm b} = \bar{\boldsymbol{y}}_1 - \bar{\boldsymbol{y}}_2$.

\section{Brief Survey of Discriminant Analysis}
\label{sec:related_work}
This section conducts a brief survey of discriminant analysis in terms of different challenges it has been tackling. We begin with its successes in discriminant dimension reduction, followed by its variants in tackling  nonlinear data, data with outliers and/or noises, data with small size, and multimodel data. Finally, we point out the multiclass LDA's limited representation ability for multiclass problems, which is also the main focus of this paper.
Readers familiar with discriminant analysis may prefer to skip this section and continue reading from Section \ref{sec:methodology}.

\subsection{Discriminant dimension reduction}
In the fields of pattern recognition and machine learning, for example, feature extraction has proven effective in mitigating complexity, improving efficiency, and enhancing classification performance \cite{yu2004efficient}. It can also be considered as subspace learning since it seeks to identify a low-dimensional representation of high-dimensional data, i.e., discovering a projection matrix using which to transform the original high-dimensional data into a low-dimensional subspace.

Fisher discriminant analysis (FDA) \cite{fisher1936use} is one of the most commonly used techniques for linear discriminant supervised dimensionality reduction. It requires an embedding transformation that maximises the between-class scatter and minimises the within-class scatter, tackling binary problems. In 1948, Rao et al. \cite{rao1948utilization} extended it to multiclass problems by solving a generalised eigenvalue problem, which becomes the most popular (classic) LDA in use presently. 
To handle multiclass problems with nonlinear challenges, variants with kernelised strategies have been developed \cite{mika1999fisher,baudat2000generalized}. The performance of the kernelised techniques is highly dependent on the choice of the kernel function family and parameters for supervised dimensionality reduction.

\subsection{Data with outliers and noises}
Classic-LDA utilises the $\ell_2$-norm distance, which may be sensitive to outliers and noises. By weighing the relative contributions of pairwise terms, weighted pairwise Fisher criteria minimising the effect of some dominant terms on the final criterion were proposed \cite{loog2001multiclass}. They are more robust compared to the classic-LDA and are capable of limiting the impact of outlier classes on the final linear dimension reduction transformation, although there is no guarantee that they would always lead to a classification rate increase since numerous estimates were required to obtain a computationally simple solution. 

Li et al. \cite{li2010linear} proposed to utilise a rotationally invariant $\ell_1$-norm to measure the two scatter matrices for discriminant projection learning. Using a weighting parameter, the relative importance of the two scatter matrices is balanced. Its application is limited due to its difficulty in identifying the optimal weighting value for varied tasks. There are many LDA variants based on the $\ell_1$-norm, e.g. \cite{wang2013fisher,zhong2013linear}, whereas obtaining the global solution is challenging. 
For example, an iterative technique was used in \cite{wang2013fisher} to obtain a local solution with each projection vector obtained repeatedly; a set of local optimal projection vectors was learned by a more robust version proposed in \cite{zhong2013linear} using a greedy strategy. 
Note that the local optimal projection vectors obtained do not always optimise their objective function. To improve the performance, a non-greedy variant of the iterative technique was developed in \cite{liu2016non}. 

\subsection{Data with small size}
For data with a small sample size, occurring when the number of data samples is much less than the number of the features, i.e., $N \ll M$, the performance of LDA methods may be degraded. 
The work in \cite{martinez2001pca} showed that principal component analysis (PCA) outperforms LDA when the amount of training data is limited. A great deal of discriminant analysis research has focused on developing methods that can better address the small sample size challenge, e.g., the methods utilising regularisation terms in the scatter matrix \cite{friedman1989regularized,lu2005regularization}, applying PCA for LDA (where PCA was used to obtain an intermediary subspace) \cite{belhumeur1997eigenfaces,liu1992efficient}, and using null space projections (to eliminate the null space of the within-class matrix) \cite{liu2004null,yang2008null,chen2000new}. 

\subsection{Multimodal data}
For multimodal data, each data class may contain several different modalities and each modality may form a separate cluster. LDA might produce unreliable results for multimodal data since it attempts to separate class means as much as possible \cite{fukunaga2013introduction}. Many LDA variants were proposed to address the multimodal data challenge.

An eigenvector-based linear dimensionality reduction approach (a multiclass extension of the technique in \cite{loog2002non}) for multiclass data with heteroskedasticity was proposed in \cite{duin2004linear}. 
Attention was paid to heteroskedasticity data by generalising the scatter between classes using a Chernoff distance metric and using the separation information from the class mean and class covariance matrix. Moreover, the mixture discriminant analysis (MDA) was proposed in \cite{hastie1996discriminant} and a combination of Gaussians to represent MDA subclasses was proposed in  \cite{zhu2006subclass}. Due to the difficulty in determining the number of mixing components, an iterative process was proposed by Gkalelis et al. \cite{gkalelis2012mixture} to estimate the number of mixing components in a Gaussian mixture model.

Many other LDA variants tackled multimodal data by examining manifold (or Laplacian graph) that depicts local structures (which are often more relevant than global structures when there are insufficient training samples for discriminant analysis). For example, the local FDA (LFDA) and local sensitive discriminant analysis (LSDA) were proposed in \cite{sugiyama2007dimensionality} and  \cite{cai2007locality}, respectively. LFDA combines FDA with a locality preserving projection that can efficiently handle multimodal data; LSDA determines a projection that optimises the distance between data points belonging to various classes in each local region. It was highlighted in \cite{zhou2016manifold} that Laplacian-based approaches only take pairwise differences into account and disregard regional consistency, and suggested manifold partition discriminant analysis. It seeks to identify a linear embedding space in which intra-class similarity is obtained along a direction consistent with local variations in the structure of the data stream and nearby data belonging to different classes are kept separate.

Non-parametric discriminant analysis (NDA) is another strategy for multimodal data. The work in \cite{fukunaga1983nonparametric} suggested a non-parametric extension based on the frequently used scatter matrix for the binary scenario. It is also regarded as a $k$-nearest neighbour version of the non-parametric valley finding technique. A non-parametric feature analysis (NFA) was presented in \cite{li2009nonparametric} based on the null space of the intra-class scatter matrix. The work in \cite{yang2011classifiers} focused on locally averaged nearest neighbour discriminant analysis and proposed a method by combining with classifiers. A different local discriminator was created in \cite{zhu2022neighborhood} to directly specify the scattering matrix across the neighbourhood. The discriminator does not require an independently and identically distributed assumption, and the neighbourhood can be viewed naturally as the smallest subclass, which is easier to acquire than subclasses without a clustering technique.

\subsection{Limitation}
It is understood that LDA methods for the multiclass case are only able to obtain $(C-1)$ discriminant directions (recall that $C$ is the number of classes for a given problem). Foley and Sammon in \cite{foley1975optimal} addressed this limitation for the binary (i.e., two-class) problem in 1975. They proposed an algorithm to derive a collection of orthogonal discriminant directions for the binary problem, in which the criteria used for selecting each discriminant direction are directly related to the discriminating potential of each direction. The number of obtained orthogonal discriminant directions  can be as large as $M$, i.e., the maximum number of orthogonal directions in $\mathbb{R}^M$. In \cite{fabiyi2021folded}, folded LDA was proposed in order to extract more than $(C-1)$ features from hyperspectral images. However, the number of the obtained discriminant directions (which are in a different dimensional space from the given samples and are not optimal) by the folded LDA is still restricted by the rank of its between-class variance matrix as what the multiclass LDA suffers. 

In this paper, we focus on the very challenging/general problem, i.e., the multiclass problem, and propose our GO-LDA, which is capable of deriving a collection and maximum number of orthogonal and optimal discriminant directions in $\mathbb{R}^M$.

\section{Linear Discriminant Analysis} \label{sec:methodology}
This section briefly presents the mathematical foundations of LDA for both the two-class and multiclass cases and the rationale for their limited discriminant ability.

Let $\boldsymbol{v}\in\mathbb{R}^M$ be an LDA projection vector. Following the Fisher criterion,  $\boldsymbol{v}$ maximises
\begin{equation}
{\cal R}(\boldsymbol{v}) =\frac{\boldsymbol{v}^{\top} \boldsymbol{S}_{\rm B} \boldsymbol{v}}{\boldsymbol{v}^{\top} \boldsymbol{S}_{\rm W} \boldsymbol{v}},\label{fisher_criterion}
\end{equation}
which is also called the Fisher ratio.
Note that $\boldsymbol{S}_{\rm B}$ and $\boldsymbol{S}_{\rm W}$ defined in Eq. \eqref{SB_SW} are the between-class scatter and within-class scatter, respectively.  
Unlike PCA which is variance preserving of the whole data, LDA seeks to find directions along which discrimination of projected data is maximised. The insight from Fisher was to find a direction along which the projected means of individual classes are as far apart as possible while simultaneously data from each class are projected with minimum scatter. This is achieved by defining an objective given by  Eq. \eqref{fisher_criterion}.

\subsection{Two-class case}
We start with the binary classification case ($C = 2$) for which Foley and Sammon \cite{foley1975optimal} derived a sequential construction of mutually orthogonal vectors that maximise the Fisher criterion ({\it cf.} Eq. \eqref{fisher_criterion}), i.e.,
\begin{equation}
\tilde{{\cal R}}(\boldsymbol{d}) = \frac{\boldsymbol{d}^{\top} \tilde{\boldsymbol{S}}_{\rm B} \boldsymbol{d}}{\boldsymbol{d}^{\top} {\boldsymbol{S}}_{\rm W} \boldsymbol{d}}.
\label{Eqn:da-fc}
\end{equation}
Note that $\tilde{{\cal R}}(\boldsymbol{d})$ is independent of the magnitude of $\boldsymbol{d}$.

The first discriminant direction, say $\boldsymbol{d}_1$, is founded by maximising $\tilde{{\cal R}}(\cdot)$, and then we have
\begin{equation} \label{eqn:da-d1}
    \boldsymbol{d}_{1} = \alpha_1 {{\boldsymbol{S}}_{\rm W}}^{-1}\boldsymbol{s}_{\rm b},
\end{equation}
where $\alpha_1$ (i.e., $\alpha_{1}^{2}=(\boldsymbol{s}_{\rm b}^{\top}[{\boldsymbol{S}}_{\rm W}^{-1}]^{2} \boldsymbol{s}_{\rm b})^{-1}$) is the normalising constant such that $\|\boldsymbol{d}_{1}\|_2 = 1$  and recall that $\boldsymbol{s}_{\rm b}$ defined in Eq. \eqref{eqn:bi-intra-inter} is the difference of the means of the two classes. Note that the discriminant direction obtained in Eq. \eqref{eqn:da-d1} is also the classic-LDA for the binary case.

The second discriminant direction $\boldsymbol{d}_{2}$ is required to maximise $\tilde{{\cal R}}(\cdot)$ in Eq. \eqref{Eqn:da-fc} and be orthogonal to $\boldsymbol{d}_{1}$, which can be derived by using the Lagrange multipliers, i.e., finding the stationary points from the Lagrangian function
\begin{equation}
\tilde{{\cal R}}(\boldsymbol{d}_2) - \lambda[\boldsymbol{d}_{2}^{\top} \boldsymbol{d}_{1}],
\label{Eqn:d2}
\end{equation}
where $\lambda$ is the Lagrange multiplier. We can then obtain
\begin{equation}
\boldsymbol{d}_{2} = 
\alpha_2 \left({\boldsymbol{S}}_{\rm W}^{-1} - \frac{\boldsymbol{s}_{\rm b}^{\top}({{\boldsymbol{S}}_{\rm W}}^{-1})^{2} \boldsymbol{\boldsymbol{s}_{\rm b}}}{\boldsymbol{s}_{\rm b}^{\top}({{\boldsymbol{S}}_{\rm W}}^{-1})^{3} \boldsymbol{s}_{\rm b}} ({{\boldsymbol{S}}_{\rm W}}^{-1})^{2}\right) {\boldsymbol{s}_{\rm b}},
\end{equation}
where $\alpha_2$ is the normalising constant such that $\|\boldsymbol{d}_{2}\|_2 = 1$;
see Appendix \ref{Appendix-dd} for the detailed derivation.

The above procedure can be extended to derive any number of discriminant directions recursively as follows. The $n$-th discriminant direction $\boldsymbol{d}_{n}$ is required to maximise $\tilde{{\cal R}}(\cdot)$ in Eq. \eqref{Eqn:da-fc} and be orthogonal to $\boldsymbol{d}_{k}, k = 1,2, \cdots, n-1$.
It can be shown that
\begin{equation}
\label{eqn:dn}
\boldsymbol{d}_{n} = \alpha_n
{{\boldsymbol{S}}_{\rm W}}^{-1}\!\left\{\boldsymbol{s}_{\rm b}-\left[\boldsymbol{d}_{1} \cdots \boldsymbol{d}_{n-1}\right] \boldsymbol{S}_{n-1}^{-1}\left[\begin{array}{c}
1/\alpha_{1} \\
0 \\
\vdots \\
\vdots \\
0
\end{array}\right]\!\right\},
\end{equation}
where $\alpha_n$ is the normalising constant such that $\|\boldsymbol{d}_{n}\|_2 = 1$ and matrix $\boldsymbol{S}_{n-1} \in \mathbb{R}^{(n-1)\times (n-1)}$ whose $(i, j)$ entries are defined as 
\begin{equation} \label{eqn-da-sn}
\boldsymbol{d}_{i}^\top {{\boldsymbol{S}}_{\rm W}}^{-1} \boldsymbol{d}_{j}, \ \  1\le i, j \le n-1.
\end{equation} 

It is worth highlighting that the way of obtaining $\boldsymbol{d}_{n}$ in Eq. \eqref{eqn:dn} only works for the binary case rather than the multiclass case, since the representation of $\boldsymbol{s}_{\rm b}$ defined in Eq. \eqref{eqn:bi-intra-inter} is only for the binary case and no such a representation for the multiclass case.

After obtaining a collection of $n$ ($n \le M$) mutually orthogonal discriminant directions, i.e., $\boldsymbol{d}_{1}, \boldsymbol{d}_{2}, \cdots, \boldsymbol{d}_{n} \in \mathbb{R}^M$, a transformation matrix $\boldsymbol{W}  =\left(\boldsymbol{d}_{1},\ldots, \boldsymbol{d}_{n}\right) \in \mathbb{R}^{M \times n}$ can be formed, which will be used to transform $\forall \boldsymbol{y} \in \mathbb{R}^{M}$ to a vector in the low-dimensional space, i.e., $\boldsymbol{W}^\top\boldsymbol{y} \in \mathbb{R}^{n}$.

\subsection{Multiclass case} \label{subsect:c-lda-multic}
Multiclass LDA seeks discriminant directions  maximising ${\cal R}(\cdot)$ in Eq. \eqref{fisher_criterion}, which, utilising the Rayleigh-Ritz quotient approach\cite{meirovitch1990convergence},  can be simplified as
\begin{equation}\label{maximise}
\underset{\boldsymbol{v}}{\operatorname{max}} \   \boldsymbol{v}^{\top} \boldsymbol{S}_{\rm B} \boldsymbol{v}, \quad 
\text {s.t.}\   \boldsymbol{v}^{\top} \boldsymbol{S}_{\rm W} \boldsymbol{v}=1.
\end{equation}
The Lagrangian reads 
\begin{equation}
\mathcal{L} (\boldsymbol{v}, \lambda) =\boldsymbol{v}^{\top} \boldsymbol{S}_{\rm B} \boldsymbol{v} -\lambda(\boldsymbol{v}^{\top} \boldsymbol{S}_{\rm W} \boldsymbol{v}-1),
\end{equation}
where $\lambda$ is the Lagrange multiplier. 
Finding the stationary points of $\mathcal{L}$ regarding $\boldsymbol{v}$ 
yields
\begin{equation}
\begin{array}{l}
\frac{\partial \mathcal{L}}{\partial \boldsymbol{v}}=2 \boldsymbol{S}_{\rm B} \boldsymbol{v}-2 \lambda \boldsymbol{S}_{\rm W} \boldsymbol{v} \stackrel{\text { set }}{=} \mathbf{0},
\end{array}
\end{equation}
i.e.,
\begin{equation}
\boldsymbol{S}_{\rm B} \boldsymbol{v} =\lambda \boldsymbol{S}_{\rm W} \boldsymbol{v}, 
\label{Eqn:generalized}
\end{equation}
which is known as the generalised eigenvalue problem regarding $\boldsymbol{S}_{\rm B}$ and $\boldsymbol{S}_{\rm W}$. Therefore, 
the eigenvector say $\boldsymbol{v}_{1}$ corresponding to the largest non-zero eigenvalue say $\lambda_1$ of the generalised eigenvalue problem \eqref{Eqn:generalized} gives the first discriminant direction for the multiclass problem. This $\boldsymbol{v}_{1}$ also maximises the Fisher criterion ${\cal R}(\cdot)$ in Eq. \eqref{fisher_criterion}. 

Since the ranks of $\boldsymbol{S}_{\rm W}$ and $\boldsymbol{S}_{\rm B}$ are $(M-C)$ and $(C-1)$, respectively, the generalised eigenvalue problem  \eqref{Eqn:generalized} has maximum of $(C-1)$ non-zero eigenvalues. After obtaining the remaining $(C-2)$ eigenvectors, i.e., $\boldsymbol{v}_{i}, i = 2, \ldots, C-1$, corresponding to the remaining $(C-2)$ eigenvalues, i.e., $\lambda_i, i = 2, \ldots, C-1$, together with $\boldsymbol{v}_1$ forming the $(C-1)$ discriminant directions, that is the so-called classic-LDA for the multiclass problem. 

For the classic-LDA, the $(C-1)$ discriminant directions can also be obtained by conducting eigendecomposition of $(\boldsymbol{S}_{\rm W}^{-1} \boldsymbol{S}_{\rm B})$.
A small perturbation can be adopted to deal with the singularity of $\boldsymbol{S}_{\rm W}$ \cite{li2006using}, which is replaced by e.g., 
\begin{equation}
{\boldsymbol{S}}_{\rm W} \leftarrow \boldsymbol{S}_{\rm W}+\delta \boldsymbol{I},
\end{equation}
where $\boldsymbol{I}$ is the identity matrix and $\delta$ (e.g., $\delta = 5\times 10^{-3}$) is a relatively small value such that ${\boldsymbol{S}}_{\rm W}$ is non-singular and therefore invertible. 

Note that both $\boldsymbol{S}_{\rm B}$ and $\boldsymbol{S}_{\rm W}^{-1}$ are symmetric matrices. This does not imply that $(\boldsymbol{S}_{\rm W}^{-1} \boldsymbol{S}_{\rm B})$ is still symmetric.
From the formation of $\boldsymbol{S}_{\rm B}$ and $\boldsymbol{S}_{\rm W}$ in Eq. \eqref{SB_SW}, $(\boldsymbol{S}_{\rm W}^{-1} \boldsymbol{S}_{\rm B})$ in practice is asymmetric, implying that the eigenvectors (i.e., the discriminant directions of the classic-LDA) obtained by eigendecomposition of $(\boldsymbol{S}_{\rm W}^{-1} \boldsymbol{S}_{\rm B})$ may not be mutually orthogonal, see Theorem \ref{non-symmetric} below.

\begin{Theorem}\label{non-symmetric}
Let $\boldsymbol{v}_{i}, i = 1, \ldots, C-1$ be the $(C-1)$ discriminant directions obtained by solving the generalised eigenvalue problem in Eq. \eqref{Eqn:generalized}, where $\boldsymbol{v}_{i}$ is the $i$-th eigenvector corresponding to the $i$-th largest eigenvalue $\lambda_i$ for Eq. \eqref{Eqn:generalized}. 
$\forall i, j \in \{1, \cdots, C-1\}, i \neq j$, if
\begin{equation} \label{eqn:nonorth-assum}
\boldsymbol{v}_{j}^\top \boldsymbol{S}_{\rm W}^{-1} \boldsymbol{S}_{\rm B} \boldsymbol{v}_{i} 
\neq \boldsymbol{v}_{j}^\top (\boldsymbol{S}_{\rm W}^{-1} \boldsymbol{S}_{\rm B})^\top \boldsymbol{v}_{i}
\end{equation}
then
\begin{equation}
\boldsymbol{v}_{i} \not\perp \boldsymbol{v}_{j}.
\end{equation}
\end{Theorem}
\begin{Proof}
For $(\boldsymbol{v}_i, \lambda_i), 1\le i \le C-1$, we have 
\begin{equation}
\boldsymbol{S}_{\rm B} \boldsymbol{v}_i =\lambda_i \boldsymbol{S}_{\rm W} \boldsymbol{v}_i,  
\end{equation}
which yields
\begin{equation}
\boldsymbol{S}_{\rm W}^{-1}\boldsymbol{S}_{\rm B} \boldsymbol{v}_i =\lambda_i \boldsymbol{v}_i. 
\end{equation}
$\forall i, j \in \{1, \cdots, C-1\}, i \neq j$, satisfying Eq. \eqref{eqn:nonorth-assum}, we have
\begin{align}
 & \ (\lambda_i - \lambda_j) \boldsymbol{v}_j^\top \boldsymbol{v}_i  \nonumber \\
= & \ \lambda_i \boldsymbol{v}_j^\top \boldsymbol{v}_i - \lambda_j \boldsymbol{v}_i^\top \boldsymbol{v}_j \nonumber  \\
= & \ \boldsymbol{v}_j^\top \boldsymbol{S}_{\rm W}^{-1}\boldsymbol{S}_{\rm B} \boldsymbol{v}_i
- \boldsymbol{v}_i^\top \boldsymbol{S}_{\rm W}^{-1}\boldsymbol{S}_{\rm B} \boldsymbol{v}_j \nonumber   \\
= & \ \boldsymbol{v}_j^\top \boldsymbol{S}_{\rm W}^{-1}\boldsymbol{S}_{\rm B} \boldsymbol{v}_i
- \boldsymbol{v}_j^\top (\boldsymbol{S}_{\rm W}^{-1}\boldsymbol{S}_{\rm B})^\top \boldsymbol{v}_i \nonumber   \\
\neq & \ 0 \quad ({\rm using \ Eq.} \ \eqref{eqn:nonorth-assum}),
\end{align}
which implies
\begin{equation}
\boldsymbol{v}_{i} \not\perp \boldsymbol{v}_{j}.
\end{equation}
This completes the proof.
\end{Proof}

It is obvious that condition in Eq. \eqref{eqn:nonorth-assum} is quite common to be satisfied given $(\boldsymbol{S}_{\rm W}^{-1} \boldsymbol{S}_{\rm B})$ is asymmetric. Therefore, the discriminant directions, i.e., $\boldsymbol{v}_{i}, i = 1, \ldots, C-1$, obtained by the classic-LDA are generally not mutually orthogonal. Most importantly, we notice that only the first discriminant direction $\boldsymbol{v}_{1}$  maximises the Fisher criterion ${\cal R}(\boldsymbol{v})$ in Eq. \eqref{fisher_criterion}. The remainder of the discriminant directions, i.e., $\boldsymbol{v}_{i}, i = 2, \ldots, C-1$, are just the eigenvectors of the generalised eigenvalue problem \eqref{Eqn:generalized}, rather than meeting/maximising the Fisher criterion ${\cal R}(\cdot)$ in Eq. \eqref{fisher_criterion}, and therefore, are non-optimal discriminant directions. 
The discriminant ability of the discriminant directions by the classic-LDA actually dropped dramatically from the second direction $\boldsymbol{v}_{2}$. The reason is that the Fisher ratio ${\cal R}(\cdot)$ in Eq. \eqref{fisher_criterion} is equal to the generalised eigenvalue $\lambda_i$ of Eq. \eqref{Eqn:generalized} for the $i$-th discriminant direction $\boldsymbol{v}_{i}$; however, the eigenvalue $\lambda_i$ becomes very small when $i \ (\le C-1)$ is growing, indicating the dramatic drop of $\boldsymbol{v}_{i}$'s discriminant ability when $i$ is large. Our illustrated experiments in Sections \ref{sec:proposed_method} and \ref{sec:experiments} also show that the discriminant directions by the classic-LDA, i.e., $\boldsymbol{v}_{i}, i \le C-1$, indeed carry more and more limited (non-optimal) discriminant information. Our proposed GO-LDA in Section \ref{sec:proposed_method} below overcomes all of these aforementioned limitations in the classic-LDA for the multiclass problem.

\section{Proposed method}
\label{sec:proposed_method}
We first present a straightforward trial of improving the classic-LDA's performance, i.e., orthogonalising the classic-LDA's non-orthogonal discriminant directions using the Gram–Schmidt process \cite{daniel1976reorthogonalization}. 
Then we propose our GO-LDA, which can produce the maximum number of optimal and also mutually orthogonal discriminant directions in space $\mathbb{R}^M$. Afterwards, we conduct a computational complexity analysis and provide some illustrations of the discriminant ability and projections of data by GO-LDA.

\subsection{Gram-Schmidt process}\label{gram-schmidt}
The Gram–Schmidt process is a way of orthonormalizing a set of non-orthogonal vectors in an inner product space; e.g., the most commonly used Euclidean space ${\mathbb{R}}^{M}$ equipped with the standard inner product defined as $\forall \boldsymbol{x}_i, \boldsymbol{x}_j \in \mathbb{R}^{M}, \langle \boldsymbol{x}_i, \boldsymbol{x}_j \rangle = \boldsymbol{x}_i^\top \boldsymbol{x}_j$. For the mutually non-orthogonal discriminant directions, i.e., $\boldsymbol{v}_{i}, i = 1, \ldots, C-1$, obtained by the classic-LDA, 
the Gram-Schmidt process with $\tilde{\boldsymbol{{v}}}_{1}=\boldsymbol{v}_{1}$, for $i = 2, \ldots, C-1$, reads,
\begin{align}
\tilde{\boldsymbol{v}}_{i} = 
\boldsymbol{v}_{i} 
- \sum_{k = 1}^{i-1} \left\langle \boldsymbol{v}_{i}, \tilde{\boldsymbol{v}}_{k}\right\rangle \frac{\tilde{\boldsymbol{v}}_{k}}{\left\|\tilde{\boldsymbol{v}}_{k}\right\|_{2}^{2}}  \ .
\end{align}
Then $\tilde{\boldsymbol{v}}_{1} \perp \tilde{\boldsymbol{v}}_{2} \perp ... \perp \tilde{\boldsymbol{v}}_{C-1}$ (normalised) will be used as discriminant directions replacing the mutually non-orthogonal ones from the classic-LDA.

We found that orthogonalisation improves the performance of the classic-LDA to some extent, but will not change the number of discriminant directions. Moreover, the orthogonalised discriminant directions are still non-optimal (since the Fisher criterion is not maximised by $\tilde{\boldsymbol{v}}_{i}, i = 2, \ldots, C-1$). Without loss of generality, in the rest of the paper, we assume all the directions are normalised.

\subsection{Generalised optimal LDA}
This section presents our GO-LDA method, deriving as many as $M$ optimal mutually orthogonal discriminant directions in $\mathbb{R}^M$. Note that in $\mathbb{R}^M$, it is understood that only $M$ mutually orthogonal directions exist. 

Let $\boldsymbol{u}_{n}, n = 1, \ldots, M$, be the discriminant directions of GO-LDA for the multiclass problem. The first GO-LDA's discriminant direction $\boldsymbol{u}_{1}$ is defined as 
\begin{equation} \label{eqn:go-lda-d1}
\boldsymbol{u}_{1} = \boldsymbol{v}_{1}
\end{equation}
(i.e., the same as that of the classic-LDA), which is the eigenvector corresponding to the largest eigenvalue of the generalised eigenvalue problem ({\it cf.} Eq. \eqref{Eqn:generalized}), maximising the Fisher criterion ${\cal R}(\cdot)$ in Eq. \eqref{fisher_criterion}. We require the second GO-LDA's discriminant direction $\boldsymbol{u}_{2}$ to maximise the Fisher criterion ${\cal R}(\cdot)$ in Eq. \eqref{fisher_criterion} and to be orthogonal to the first discriminant direction $\boldsymbol{u}_{1}$. Then we derive that $\boldsymbol{u}_{2}$ can be found by solving the generalised eigenvalue problem given in Eq. \eqref{eqn:go-lda-u2-gep}, see Theorem \ref{thm:go-lda-u2} below.

\begin{theorem} \label{thm:go-lda-u2}
Let $\boldsymbol{u}_{2}\in \mathbb{R}^M$  maximise the Fisher criterion ${\cal R}(\cdot)$ in Eq. \eqref{fisher_criterion} and be orthogonal to $\boldsymbol{u}_{1}$ in Eq. \eqref{eqn:go-lda-d1}, i.e., $\boldsymbol{u}_{2}$ is the solution of the following problem
\begin{equation} \label{eqn:p-go-lda-u2}
\underset{\boldsymbol{u}}{\operatorname{max}} \
{\cal R}(\boldsymbol{u}) =\frac{\boldsymbol{u}^{\top} \boldsymbol{S}_{\rm B} \boldsymbol{u}}{\boldsymbol{u}^{\top} \boldsymbol{S}_{\rm W} \boldsymbol{u}},
\quad {\rm s.t.} \ 
\boldsymbol{u} \perp \boldsymbol{u}_{1} .
\end{equation}
Then $\boldsymbol{u}_{2}$ is the eigenvector corresponding to the largest eigenvalue of the  generalised eigenvalue problem 
\begin{equation}  \label{eqn:go-lda-u2-gep}
\left(\boldsymbol{S}_{\rm B}-\boldsymbol{k}_{1}\right) \boldsymbol{u}={\mu} \boldsymbol{S}_{\rm W} \boldsymbol{u},
\end{equation}
where 
\begin{equation} \label{eqn:go-lda-k1}
\boldsymbol{k}_{1} = 
\frac{\boldsymbol{u}_{1}^{\top} \boldsymbol{S}_{\rm W}^{-1}  \boldsymbol{S}_{\rm B} \boldsymbol{u}_{1}}{\boldsymbol{u}_{1}^{\top} \boldsymbol{S}_{\rm W}^{-1} \boldsymbol{u}_{1}}.
\end{equation}
\end{theorem}
\begin{Proof}
The maximisation problem \eqref{eqn:p-go-lda-u2} can be derived by finding the stationary points from the Lagrangian function
\begin{equation}
\hat{{\cal L}}(\boldsymbol{u}, \beta)
=
\frac{\boldsymbol{u}^{\top} \boldsymbol{S}_{\rm B} \boldsymbol{u}}{\boldsymbol{u}^{\top} \boldsymbol{S}_{\rm W} \boldsymbol{u}} - \beta \boldsymbol{u}^{\top} \boldsymbol{u}_{1},
\label{d2_criterion}
\end{equation}
where $\beta$ is the Lagrange multiplier.
This means its partial derivation with respect to $\boldsymbol{u}$ should be zero, i.e., 
\begin{equation} \label{eqn:lp-go-lda-u2}
\frac{2\boldsymbol{S}_{\rm B}\boldsymbol{u}\boldsymbol{u}^{\top} \boldsymbol{S}_{\rm W}\boldsymbol{u} - 2 \boldsymbol{u}^{\top}\boldsymbol{S}_{\rm B}\boldsymbol{u} \boldsymbol{S}_{\rm W} \boldsymbol{u}}{(\boldsymbol{u}^{\top} \boldsymbol{S}_{\rm W} \boldsymbol{u})^2} - \beta \boldsymbol{u}_{1}=0.
\end{equation}
Since $\boldsymbol{u}_{1}^\top\boldsymbol{u} = 0$ and note that $\boldsymbol{u}^{\top}\boldsymbol{S}_{\rm B}\boldsymbol{u}$ is a scalar, we have 
\begin{equation} \label{eqn:lp-go-lda-u2-temp}
(\boldsymbol{u}_{1}^{\top}\boldsymbol{S}_{\rm W}^{-1}) (2 \boldsymbol{u}^{\top}\boldsymbol{S}_{\rm B}\boldsymbol{u} \boldsymbol{S}_{\rm W} \boldsymbol{u}) = 
2 \boldsymbol{u}^{\top}\boldsymbol{S}_{\rm B}\boldsymbol{u} (\boldsymbol{u}_{1}^{\top} \boldsymbol{u})
= 0.
\end{equation}
Multiplying $\boldsymbol{u}_{1}^{\top}\boldsymbol{S}_{\rm W}^{-1}$ on both sides of Eq. \eqref{eqn:lp-go-lda-u2} and using Eq. \eqref{eqn:lp-go-lda-u2-temp} yield
\begin{equation}
\frac{2\boldsymbol{u}_{1}^{\top} \boldsymbol{S}_{\rm W}^{-1} \boldsymbol{S}_{\rm B} \boldsymbol{u}}{\boldsymbol{u}^{\top} \boldsymbol{S}_{\rm W} \boldsymbol{u}}- \beta\boldsymbol{u}_{1}^{\top} \boldsymbol{S}_{\rm W}^{-1} \boldsymbol{u}_{1}=0.
\end{equation}
Then we have 
\begin{equation}
\beta = \frac{2 \boldsymbol{u}_{1}^{\top} \boldsymbol{S}_{\rm W}^{-1} \boldsymbol{S}_{\rm B} \boldsymbol{u}}{(\boldsymbol{u}^{\top} \boldsymbol{S}_{\rm W} \boldsymbol{u})\boldsymbol{u}_{1}^{\top} \boldsymbol{S}_{\rm W}^{-1} \boldsymbol{u}_{1}}.
\end{equation}

Substituting the representation of $\beta$ above into Eq. \eqref{eqn:lp-go-lda-u2} and then multiplying $\boldsymbol{u}^{\top} \boldsymbol{S}_{\rm W} \boldsymbol{u}/2$ on both sides, we have
\begin{equation} \label{eqn:lp-go-lda-u2-temp2}
\boldsymbol{S}_{\rm B}\boldsymbol{u}
-
\frac{\boldsymbol{u}^{\top}\boldsymbol{S}_{\rm B}\boldsymbol{u}}{\boldsymbol{u}^{\top} \boldsymbol{S}_{\rm W} \boldsymbol{u}}  \boldsymbol{S}_{\rm W} \boldsymbol{u} 
-  
\frac{\boldsymbol{u}_{1}^{\top} \boldsymbol{S}_{\rm W}^{-1}  \boldsymbol{S}_{\rm B} \boldsymbol{u}_{1}}{\boldsymbol{u}_{1}^{\top} \boldsymbol{S}_{\rm W}^{-1} \boldsymbol{u}_{1}}
\boldsymbol{u}=0,
\end{equation}
which can be rewritten as 
\begin{equation} 
\left(\boldsymbol{S}_{\rm B}
-  
\frac{\boldsymbol{u}_{1}^{\top} \boldsymbol{S}_{\rm W}^{-1}  \boldsymbol{S}_{\rm B} \boldsymbol{u}_{1} }{\boldsymbol{u}_{1}^{\top} \boldsymbol{S}_{\rm W}^{-1} \boldsymbol{u}_{1}} \right)
\boldsymbol{u}
=
\frac{\boldsymbol{u}^{\top}\boldsymbol{S}_{\rm B}\boldsymbol{u}}{\boldsymbol{u}^{\top} \boldsymbol{S}_{\rm W} \boldsymbol{u}}  \boldsymbol{S}_{\rm W} \boldsymbol{u} ,
\end{equation}
i.e. (using the definition of $\boldsymbol{k}_{1}$ in Eq. \eqref{eqn:go-lda-k1}), 
\begin{equation} \label{eqn:lp-go-lda-u2-temp3}
\left(\boldsymbol{S}_{\rm B}
-  
\boldsymbol{k}_{1} \right)
\boldsymbol{u}
=
\frac{\boldsymbol{u}^{\top}\boldsymbol{S}_{\rm B}\boldsymbol{u}}{\boldsymbol{u}^{\top} \boldsymbol{S}_{\rm W} \boldsymbol{u}}  \boldsymbol{S}_{\rm W} \boldsymbol{u} .
\end{equation}
Since $\boldsymbol{u}_2$ satisfies the problem in \eqref{eqn:p-go-lda-u2}, it is also a solution of Eq. \eqref{eqn:lp-go-lda-u2-temp3} above, i.e., 
\begin{equation} 
\left(\boldsymbol{S}_{\rm B}
-  
\boldsymbol{k}_{1} \right)
\boldsymbol{u}_2
=
\frac{\boldsymbol{u}_2^{\top}\boldsymbol{S}_{\rm B}\boldsymbol{u}_2}{\boldsymbol{u}_2^{\top} \boldsymbol{S}_{\rm W} \boldsymbol{u}_2}  \boldsymbol{S}_{\rm W} \boldsymbol{u}_2 .
\end{equation}
This means $\boldsymbol{u}_2$ is an eigenvector of the eigenvalue
\begin{equation}
\mu_1^{(2)}
=
\frac{\boldsymbol{u}_2^{\top}\boldsymbol{S}_{\rm B}\boldsymbol{u}_2}{\boldsymbol{u}_2^{\top} \boldsymbol{S}_{\rm W} \boldsymbol{u}_2}
\end{equation}
for the generalised eigenvalue problem in \eqref{eqn:go-lda-u2-gep},
with $\boldsymbol{k}_{1}$ defined in \eqref{eqn:go-lda-k1}. Since $\boldsymbol{u}_2$ maximises 
${\cal R}(\boldsymbol{u})$ in \eqref{eqn:p-go-lda-u2}, indicating $\mu_1^{(2)}$ is the largest eigenvalue of the generalised eigenvalue problem in \eqref{eqn:go-lda-u2-gep}. This completes the proof.
\end{Proof}

Theorem \ref{thm:go-lda-u2} shows that the second GO-LDA's discriminant direction $\boldsymbol{u}_{2}$ (also orthogonal to $\boldsymbol{u}_{1}$) is the eigenvector corresponding to the largest eigenvalue of the generalised eigenvalue problem given in Eq. \eqref{eqn:go-lda-u2-gep}, which is different from the generalised eigenvalue problem in Eq. \eqref{Eqn:generalized} used in the classic-LDA. For the difference between our GO-LDA and the classic-LDA in deriving their second discriminant direction, except for solving different generalised eigenvalue problems, GO-LDA takes the eigenvector corresponding to the largest eigenvalue while the classic-LDA takes the eigen- vector corresponding to the second largest eigenvalue of their generalised eigenvalue problems. It also indicates that the second discriminant direction of GO-LDA is optimal whereas that of the classic-LDA is not.

\begin{algorithm*}[ht]
\caption{GO-LDA: Generalised Optimal LDA}\label{alg:go-lda}
\begin{algorithmic}
\STATE \textbf{Input:} Data $\boldsymbol{Y}\in\mathbb{R}^{N\times M}$, number of classes $C$, and $K \le M$ number of  discriminant directions.
\STATE \textbf{Output:}
Discriminant directions $\{\boldsymbol{u}_{n}\}_{n=1}^K$.
\STATE \hspace{0.3cm} {Compute} 
${\boldsymbol{S}}_{\rm W}$ and $\boldsymbol{S}_{\rm B}$ in Eq. \eqref{SB_SW};
\STATE \hspace{0.3cm} {Compute} $\boldsymbol{u}_{1}$ in Eq. \eqref{eqn:go-lda-d1} and normalise it;
\STATE \hspace{0.3cm} $n$ = 2;
\STATE \hspace{0.3cm} \textbf{for} {$n \le K$} \textbf{do}
\STATE \hspace{0.8cm} {Form} the
matrices $\boldsymbol{U}_{n-1}, \boldsymbol{B}_{n-1}$ and $\boldsymbol{T}_{n-1}$ using the definitions in \eqref{eqn:go-lda-U}, \eqref{eqn:go-lda-B} and \eqref{eqn:go-lda-T}, respectively.
\STATE \hspace{0.8cm} {Form}
the generalised eigenvalue problem \eqref{eqn:go-lda-ui-gep}.
\STATE \hspace{0.8cm} {Compute} 
the eigenvector $\boldsymbol{u}_{n}$ corresponding to the largest eigenvalue of the problem \eqref{eqn:go-lda-ui-gep} and normalise it;
\STATE \hspace{0.8cm} ${n}={n}+1$;
\STATE \hspace{0.3cm} \textbf{end for}
\STATE  \textbf{Return} $\{\boldsymbol{u}_{n}\}_{n=1}^K$
\end{algorithmic}
\end{algorithm*}

Now we give the derivation of all the discriminant directions of GO-LDA for the multiclass problem, i.e., $\boldsymbol{u}_{n}, n = 2, \ldots, M$, where each $\boldsymbol{u}_{n}\in \mathbb{R}^M$ maximises the Fisher criterion ${\cal R}(\cdot)$ in Eq. \eqref{fisher_criterion} and is orthogonal to the discriminant directions $\boldsymbol{u}_{i}, i = 1, \ldots, n-1$. Theorem \ref{thm:go-lda-ui} below
shows that each $\boldsymbol{u}_{n}, 2 \le n \le M$, can be derived by solving a different generalised eigenvalue problem given in Eq. \eqref{eqn:go-lda-ui-gep}.

\begin{theorem} \label{thm:go-lda-ui}
Let $\boldsymbol{u}_{n}\in \mathbb{R}^M, 2\le n \le M$  maximise the Fisher criterion ${\cal R}(\cdot)$ in Eq. \eqref{fisher_criterion} and be orthogonal to the GO-LDA's discriminant directions $\boldsymbol{u}_{i}, i = 1, \ldots, n-1$, i.e., $\boldsymbol{u}_{n}$ is the solution of the following problem
\begin{align} \label{eqn:p-go-lda-ui}
\begin{split}
& \underset{\boldsymbol{u}}{\operatorname{max}} \
 {\cal R}(\boldsymbol{u}) =\frac{\boldsymbol{u}^{\top} \boldsymbol{S}_{\rm B} \boldsymbol{u}}{\boldsymbol{u}^{\top} \boldsymbol{S}_{\rm W} \boldsymbol{u}}, \\
& \ \ {\rm s.t.} \ 
\boldsymbol{u} \perp \boldsymbol{u}_{1} \perp ... \perp \boldsymbol{u}_{n-1}.
\end{split}
\end{align}
Then $\boldsymbol{u}_{n}$ is the eigenvector corresponding to the largest eigenvalue of the  generalised eigenvalue problem
\begin{equation} \label{eqn:go-lda-ui-gep}
\left(\boldsymbol{S}_{\rm B}-\boldsymbol{U}_{n-1} \boldsymbol{T}_{n-1}^{-1} \boldsymbol{B}_{n-1}\right)\boldsymbol{u}
=
\mu \boldsymbol{S}_{\rm W} \boldsymbol{u},
\end{equation}
where matrices
\begin{equation} \label{eqn:go-lda-U}
\boldsymbol{U}_{n-1} 
= (\boldsymbol{u}_{1}\cdots\boldsymbol{u}_{n-1}), 
\end{equation}
\begin{equation} \label{eqn:go-lda-B}
\boldsymbol{B}_{n-1} = 
\left(\begin{array}{c}
\boldsymbol{u}_{1}^{\top} \boldsymbol{S}_{\mathrm{W}}^{-1} \boldsymbol{S}_{\mathrm{B}}\vspace{1ex} \\
\boldsymbol{u}_{2}^{\top} \boldsymbol{S}_{\mathrm{W}}^{-1} \boldsymbol{S}_{\mathrm{B}} \\
\vdots \\
\boldsymbol{u}_{n-1}^{\top} \boldsymbol{S}_{\mathrm{W}}^{-1} \boldsymbol{S}_{\mathrm{B}}
\end{array}\right),
\end{equation}
and $\boldsymbol{T}_{n-1} \in \mathbb{R}^{(n-1)\times (n-1)}$ whose $(i, j)$ entries are defined as 
\begin{equation} \label{eqn:go-lda-T}
\boldsymbol{u}_{i}^\top {{\boldsymbol{S}}_{\rm W}}^{-1} \boldsymbol{u}_{j}, \ \  1\le i, j \le n-1.
\end{equation} 
\end{theorem}

\begin{Proof}
The maximisation problem \eqref{eqn:p-go-lda-ui} can be derived by finding the stationary points from the Lagrangian function
\begin{equation}
\bar{{\cal L}}(\boldsymbol{u}, \boldsymbol{\beta})
=
\frac{\boldsymbol{u}^{\top} \boldsymbol{S}_{\rm B} \boldsymbol{u}}{\boldsymbol{u}^{\top} \boldsymbol{S}_{\rm W} \boldsymbol{u}}
-
\sum_{i=1}^{n-1}
\beta_{i} \boldsymbol{u}^{\top} \boldsymbol{u}_{i},
\label{criterion_for_dn}
\end{equation}
where $\boldsymbol{\beta} = (\beta_1, \cdots, \beta_{n-1})^\top$ is the Lagrange multiplier.
This means its partial derivation regarding $\boldsymbol{u}$ should be zero, i.e., 
\begin{equation} \label{eqn:lp-go-lda}
\frac{2\boldsymbol{S}_{\rm B}\boldsymbol{u}\boldsymbol{u}^{\top} \boldsymbol{S}_{\rm W}\boldsymbol{u}}{(\boldsymbol{u}^{\top} \boldsymbol{S}_{\rm W} \boldsymbol{u})^2} 
-
\frac{2\boldsymbol{u}^{\top}\boldsymbol{S}_{\rm B}\boldsymbol{u} \boldsymbol{S}_{\rm W} \boldsymbol{u}}{(\boldsymbol{u}^{\top} \boldsymbol{S}_{\rm W} \boldsymbol{u})^2} 
- 
\sum_{i=1}^{n-1}
\beta_{i} \boldsymbol{u}_{i}
=
0,
\end{equation}
which is equivalent to (using $\boldsymbol{U}_{n-1}$ defined in Eq. \eqref{eqn:go-lda-U})
\begin{equation} \label{eqn:lp-go-lda-ui}
\frac{2\boldsymbol{S}_{\rm B}\boldsymbol{u}\boldsymbol{u}^{\top} \boldsymbol{S}_{\rm W}\boldsymbol{u}}{(\boldsymbol{u}^{\top} \boldsymbol{S}_{\rm W} \boldsymbol{u})^2} 
-
\frac{2\boldsymbol{u}^{\top}\boldsymbol{S}_{\rm B}\boldsymbol{u} \boldsymbol{S}_{\rm W} \boldsymbol{u}}{(\boldsymbol{u}^{\top} \boldsymbol{S}_{\rm W} \boldsymbol{u})^2} 
- 
\boldsymbol{U}_{n-1} 
\boldsymbol{\beta}
=
0.
\end{equation}
Since $\boldsymbol{u}_{k}^\top\boldsymbol{u} = 0, k = 1, \ldots, n-1$, multiplying $\boldsymbol{u}_{k}^{\top}\boldsymbol{S}_{\rm W}^{-1},  k = 1, \ldots, n-1$, individually on both sides of Eq. \eqref{eqn:lp-go-lda-ui} and using Eq. \eqref{eqn:lp-go-lda-u2-temp} yield 
\begin{equation}
\alpha \boldsymbol{u}_{k}^{\top} \boldsymbol{S}_{\mathrm{W}}^{-1} \boldsymbol{S}_{\mathrm{B}} \boldsymbol{u}
-
\sum_{i=1}^{n-1}
\beta_{i} \boldsymbol{u}_{k}^{\top} \boldsymbol{S}_{\mathrm{W}}^{-1} \boldsymbol{u}_{i} = 0, 
\ k = 1, \ldots, n-1,
\end{equation}
i.e.,
\begin{equation} \label{eqn:go-lda-pde}
\sum_{i=1}^{n-1}
\beta_{i} \boldsymbol{u}_{k}^{\top} \boldsymbol{S}_{\mathrm{W}}^{-1} \boldsymbol{u}_{i}
=
\alpha (\boldsymbol{u}_{k}^{\top} \boldsymbol{S}_{\mathrm{W}}^{-1} \boldsymbol{S}_{\mathrm{B}}) \boldsymbol{u}, 
\ k = 1, \ldots, n-1,
\end{equation}
where 
\begin{equation}
\alpha = 2/(\boldsymbol{u}^{\top} \boldsymbol{S}_{\mathrm{W}} \boldsymbol{u}).
\end{equation}

Using the definition of $\boldsymbol{B}_{n-1}$ and $\boldsymbol{T}_{n-1}$ in Eq. \eqref{eqn:go-lda-B} and Eq. \eqref{eqn:go-lda-T}, respectively, Eq. \eqref{eqn:go-lda-pde} can be rewritten as
\begin{equation}
\boldsymbol{T}_{n-1} \boldsymbol{\beta}
=
\alpha \boldsymbol{B}_{n-1} \boldsymbol{u}.
\end{equation}
We then have 
\begin{equation}
\boldsymbol{\beta} = \alpha \boldsymbol{T}_{n-1}^{-1}  \boldsymbol{B}_{n-1} \boldsymbol{u}.
\end{equation}

Substituting the representation of  $\boldsymbol{\beta}$ above into Eq. \eqref{eqn:lp-go-lda-ui} and then multiplying $\boldsymbol{u}^{\top} \boldsymbol{S}_{\rm W} \boldsymbol{u}/2$ on both sides, we have
\begin{equation} \label{eqn:lp-go-lda-ui-temp2}
\boldsymbol{S}_{\rm B}\boldsymbol{u}
-
\frac{\boldsymbol{u}^{\top}\boldsymbol{S}_{\rm B}\boldsymbol{u}}{\boldsymbol{u}^{\top} \boldsymbol{S}_{\rm W} \boldsymbol{u}}  \boldsymbol{S}_{\rm W} \boldsymbol{u} 
-  
\boldsymbol{U}_{n-1}  \boldsymbol{T}_{n-1}^{-1} \boldsymbol{B}_{n-1}
\boldsymbol{u}=0,
\end{equation}
i.e.,
\begin{equation} \label{eqn:lp-go-lda-ui-temp3}
\left(\boldsymbol{S}_{\rm B}
-  
\boldsymbol{U}_{n-1} \boldsymbol{T}_{n-1}^{-1} \boldsymbol{B}_{n-1}
\right) \boldsymbol{u}
=
\frac{\boldsymbol{u}^{\top}\boldsymbol{S}_{\rm B}\boldsymbol{u}}{\boldsymbol{u}^{\top} \boldsymbol{S}_{\rm W} \boldsymbol{u}}  \boldsymbol{S}_{\rm W} \boldsymbol{u}.
\end{equation}

Since $\boldsymbol{u}_n$ satisfies the problem in \eqref{eqn:p-go-lda-ui}, it is also a solution of Eq. \eqref{eqn:lp-go-lda-ui-temp3} above, i.e., 
\begin{equation} 
\left(\boldsymbol{S}_{\rm B}
-  
\boldsymbol{U}_{n-1} \boldsymbol{T}_{n-1}^{-1} \boldsymbol{B}_{n-1}
\right) \boldsymbol{u}_n
=
\frac{\boldsymbol{u}_n^{\top}\boldsymbol{S}_{\rm B}\boldsymbol{u}_n}{\boldsymbol{u}_n^{\top} \boldsymbol{S}_{\rm W} \boldsymbol{u}_n}  \boldsymbol{S}_{\rm W} \boldsymbol{u}_n.
\end{equation}
This means $\boldsymbol{u}_n$ is an eigenvector of the eigenvalue say
\begin{equation}
\mu_1^{(n)}
=
\frac{\boldsymbol{u}_2^{\top}\boldsymbol{S}_{\rm B}\boldsymbol{u}_2}{\boldsymbol{u}_2^{\top} \boldsymbol{S}_{\rm W} \boldsymbol{u}_2}
\end{equation}
for the generalised eigenvalue problem in Eq. \eqref{eqn:go-lda-ui-gep}. Since $\boldsymbol{u}_n$ maximises ${\cal R}(\boldsymbol{u})$ in Eq. \eqref{eqn:p-go-lda-ui}, indicating $\mu_1^{(n)}$ is the largest eigenvalue of the generalised eigenvalue problem in Eq. \eqref{eqn:go-lda-ui-gep}. This completes the proof.
\end{Proof}

Theorem \ref{thm:go-lda-ui} tells us that GO-LDA's discriminant directions, i.e., $\boldsymbol{u}_{n}, n = 1, \ldots, M$, can be derived recursively with $\boldsymbol{u}_{1}$ computed in Eq. \eqref{eqn:go-lda-d1}, by computing the eigenvector corresponding to the largest eigenvalue of the generalised eigenvalue problem given in Eq. \eqref{eqn:go-lda-ui-gep} each time. Obviously, the GO-LDA's discriminant directions are optimal since they are all maximising the Fisher criterion ${\cal R}(\cdot)$ in Eq. \eqref{fisher_criterion} with the mutually orthogonal constraint. We finally remark that the ideas proposed in Theorems \ref{thm:go-lda-u2} and \ref{thm:go-lda-ui} could be exploited to enhance the performance of most, if not all, of the LDA variants (e.g. the ones surveyed in Section \ref{sec:related_work}).

The whole procedure of finding GO-LDA's discriminant directions, i.e., $\{\boldsymbol{u}_{n}\}_{n=1}^M$, is summarised in Algorithm \ref{alg:go-lda}.

\begin{figure*}[ht]
\centering
\begin{tabular} {cc}
\includegraphics[width=3.4in,height=2.2in]{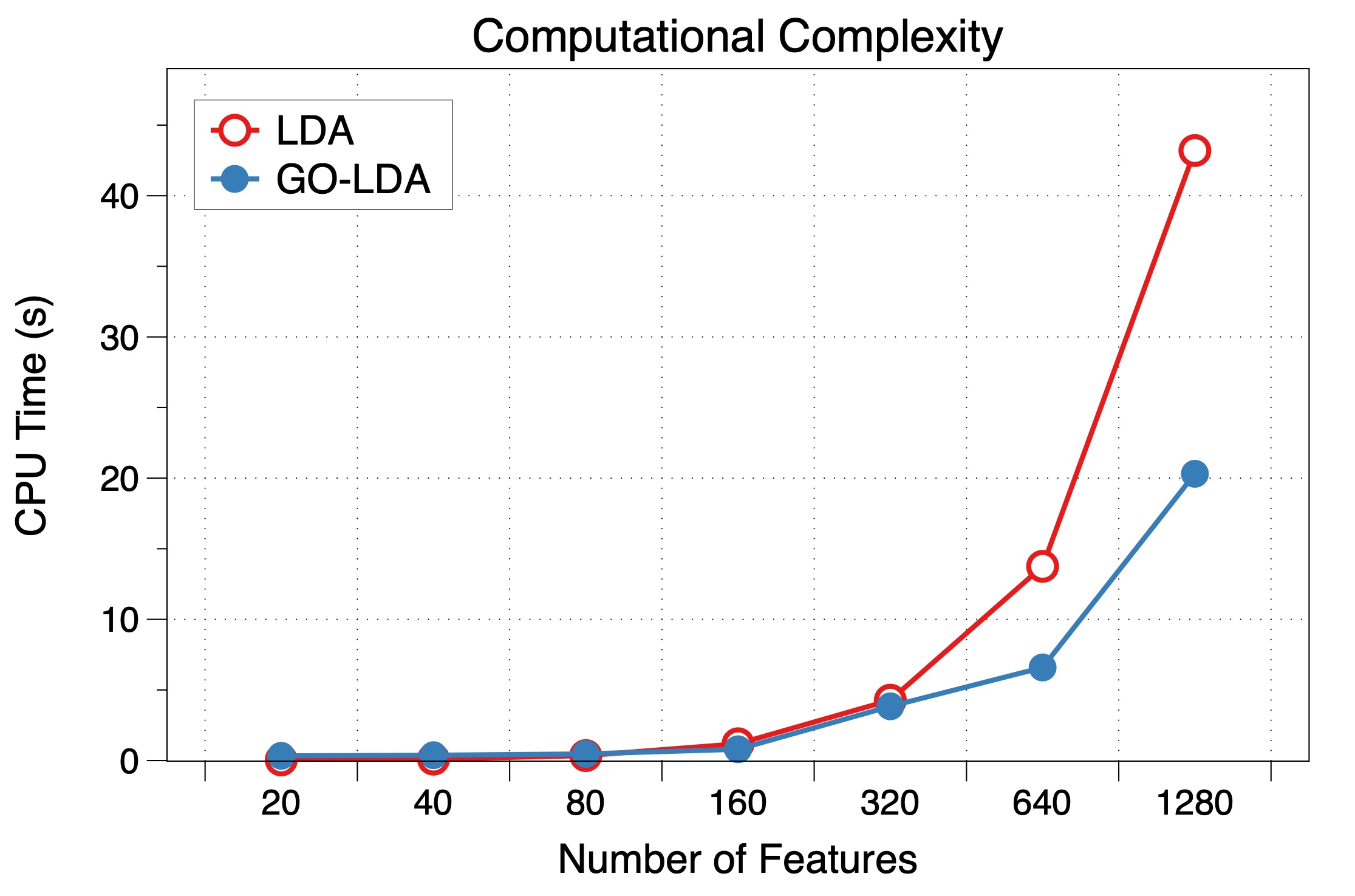}
&
\includegraphics[width=3.4in,height=2.2in]{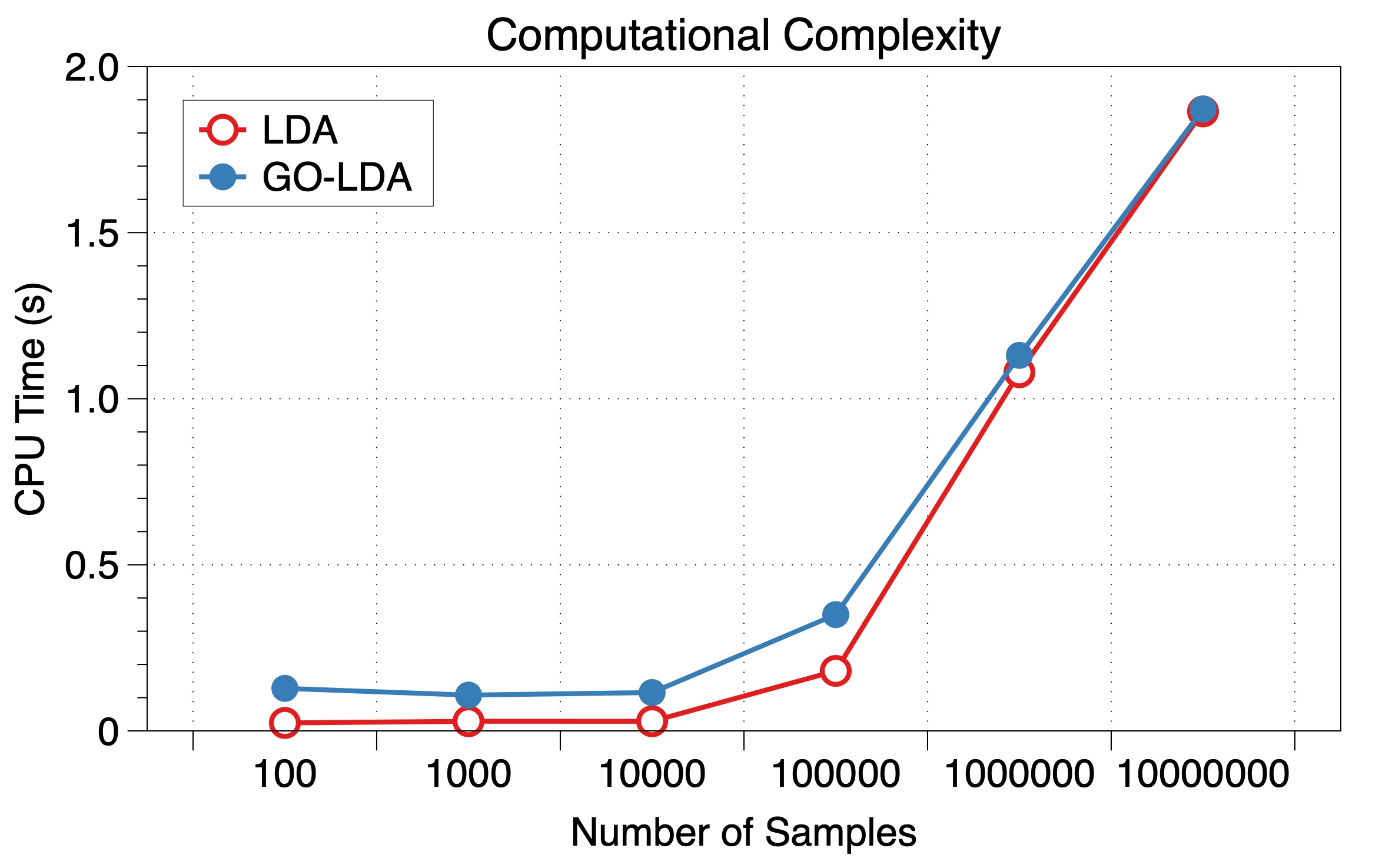} 
\end{tabular}
\caption{Computation cost comparison between the classic-LDA (referred to LDA in the plots/tables for simplicity) and GO-LDA. The left panel shows the CPU time for synthetic data sets with a fixed number of samples and an increased number of data features. The right panel shows the CPU time for synthetic data sets with an increased number of samples and a fixed number of data features. The synthetic data sets are created by using the Python built-in function {\tt make$\_$blobs} to generate isotropic Gaussian samples. The number of classes is $C = 5$, and four discriminant directions are derived for both the classic-LDA and GO-LDA. In particular, for the left panel, $N = 1,000$ with the number of data features $M$ increased from $20$ to over $1,000$; and for the right panel, $M = 10$ with the number of samples $N$ increased from 100 to over $10^{7}$. 
}
\label{fig:comp_cost}
\end{figure*}

\subsection{Computational complexity}
We now analyse GO-LDA's computational complexity and make comparison to the classic-LDA.

\subsubsection{Computation cost}
Recall that $N$ is the number of samples of the given data set and $M$ is the number of features of every sample.
The classic-LDA (presented in Section \ref{subsect:c-lda-multic}) consists of the following two  key steps. The first is to form the inter-class matrix $\boldsymbol{S}_{\rm B}$ and the intra-class matrix $\boldsymbol{S}_{\rm W}$, as specified in Eq. \eqref{SB_SW}, and the second is to solve the generalised eigenvalue problem \eqref{Eqn:generalized}.
The computational complexity of forming $\boldsymbol{S}_{\rm B}$ and $\boldsymbol{S}_{\rm W}$
is dominated by $\boldsymbol{S}_{\rm W}$, which is ${\cal O}(NM^2)$. Solving the generalised eigenvalue problem \eqref{Eqn:generalized} takes ${\cal O}(M^3)$ if e.g. we calculate $\boldsymbol{S}_W^{-1}$ first (taking ${\cal O}(M^3)$) and then conduct eigendecomposition. The computation cost of the classic-LDA is therefore 
\begin{equation} \label{eqn:lda-cost}
{\cal O}(NM^2)+ 2{\cal O}(M^3).
\end{equation}
Note that the above representation is informal since our main purpose here is for ease of comparison ({\it cf.} Eq. \eqref{eqn:golda-cost}).

Analogously, GO-LDA also firstly computes matrices $\boldsymbol{S}_{\rm B}$ and $\boldsymbol{S}_{\rm W}$, taking ${\cal O}(NM^2)$, see Algorithm \ref{alg:go-lda}. 
Forming matrices $\boldsymbol{U}_{n-1}, \boldsymbol{B}_{n-1}$ and $\boldsymbol{T}_{n-1}$ using the definitions in Eq. \eqref{eqn:go-lda-U}, \eqref{eqn:go-lda-B} and \eqref{eqn:go-lda-T}, respectively, takes ${\cal O}(M^3)$, since the computation cost is dominated by previously forming $\boldsymbol{S}_W^{-1}$, which is ${\cal O}(M^3)$.
Then, to compute $K$, $K \le M$, number of GO-LDA discriminate directions, $K$ number of generalised eigenvalue problems given in Eq. \eqref{eqn:go-lda-ui-gep} need to be solved.
Note that, for each generalised eigenvalue problem in Eq. \eqref{eqn:go-lda-ui-gep}, GO-LDA only computes one eigenvector corresponding to the largest eigenvalue (rather than the complete eigendecomposition), taking ${\cal O}(M^2)$ \cite{strang2006linear}. We then obtain the computation cost of GO-LDA, i.e., 
\begin{equation} \label{eqn:golda-cost}
{\cal O}(NM^2) + {\cal O}(M^3) + K {\cal O}(M^2).
\end{equation}

From the computation cost representations of the classic-LDA and GO-LDA respectively in \eqref{eqn:lda-cost} and \eqref{eqn:golda-cost}, we see that if $N\gg M$, then $2{\cal O}(M^3)$ in Eq. \eqref{eqn:lda-cost} and $({\cal O}(M^3) + K {\cal O}(M^2))$ in Eq. \eqref{eqn:golda-cost} will be dominated by the term ${\cal O}(NM^2)$. Moreover, $K \le M$ implies that ${\cal O}(M^3)$ and $K{\cal O}(M^2)$ are comparable. Therefore, the computation cost representations in \eqref{eqn:lda-cost} and \eqref{eqn:golda-cost} indicate that both the classic-LDA and GO-LDA have comparable computation cost, even though our GO-LDA needs to solve $K$ different generalised eigenvalue problems.

\subsubsection{Experimental demonstration}
Below we experimentally demonstrate the GO-LDA's computation cost and make comparison to the classic-LDA. The Python package {\tt np.linalg.eig} is used to perform eigendecomposition; in particular, the Python built-in function {\tt scipy.sparse.linalg.eigs} is used to calculate GO-LDA's individual discriminant directions, i.e., calculating one eigenvector corresponding to the largest eigenvalue of each generalised eigenvalue problem in Eq. \eqref{eqn:go-lda-ui-gep}.

Fig. \ref{fig:comp_cost} gives the CPU time of the classic-LDA and GO-LDA on synthetic data sets with different number of features $M$ and different number of samples $N$, see the left and right panels of Fig. \ref{fig:comp_cost}. In particular, the left panel of Fig. \ref{fig:comp_cost} shows that our GO-LDA is surprisingly even more economical than the classic-LDA when the number of features $M$ increases. This could be explained by the computation cost representations in \eqref{eqn:lda-cost} and \eqref{eqn:golda-cost}, i.e., $K {\cal O}(M^2)$ is less than ${\cal O}(M^3)$ when $K\ll M$. The right panel of Fig. \ref{fig:comp_cost} shows that, when both $M$ and $N$ are small, the CPU cost of our GO-LDA is slightly higher than that of the classic-LDA, but the difference vanishes when the number of samples $N$ increases. This could again be explained by the computation cost representations in \eqref{eqn:lda-cost} and \eqref{eqn:golda-cost}, i.e., the term ${\cal O}(NM^2)$ dominates both representations when $N$ is large. On the whole, Fig. \ref{fig:comp_cost} experimentally demonstrates that the computation cost of our GO-LDA is quite economical and is comparable to the classic-LDA.

\begin{figure}[ht]
\centering
\includegraphics[width=3.2in]{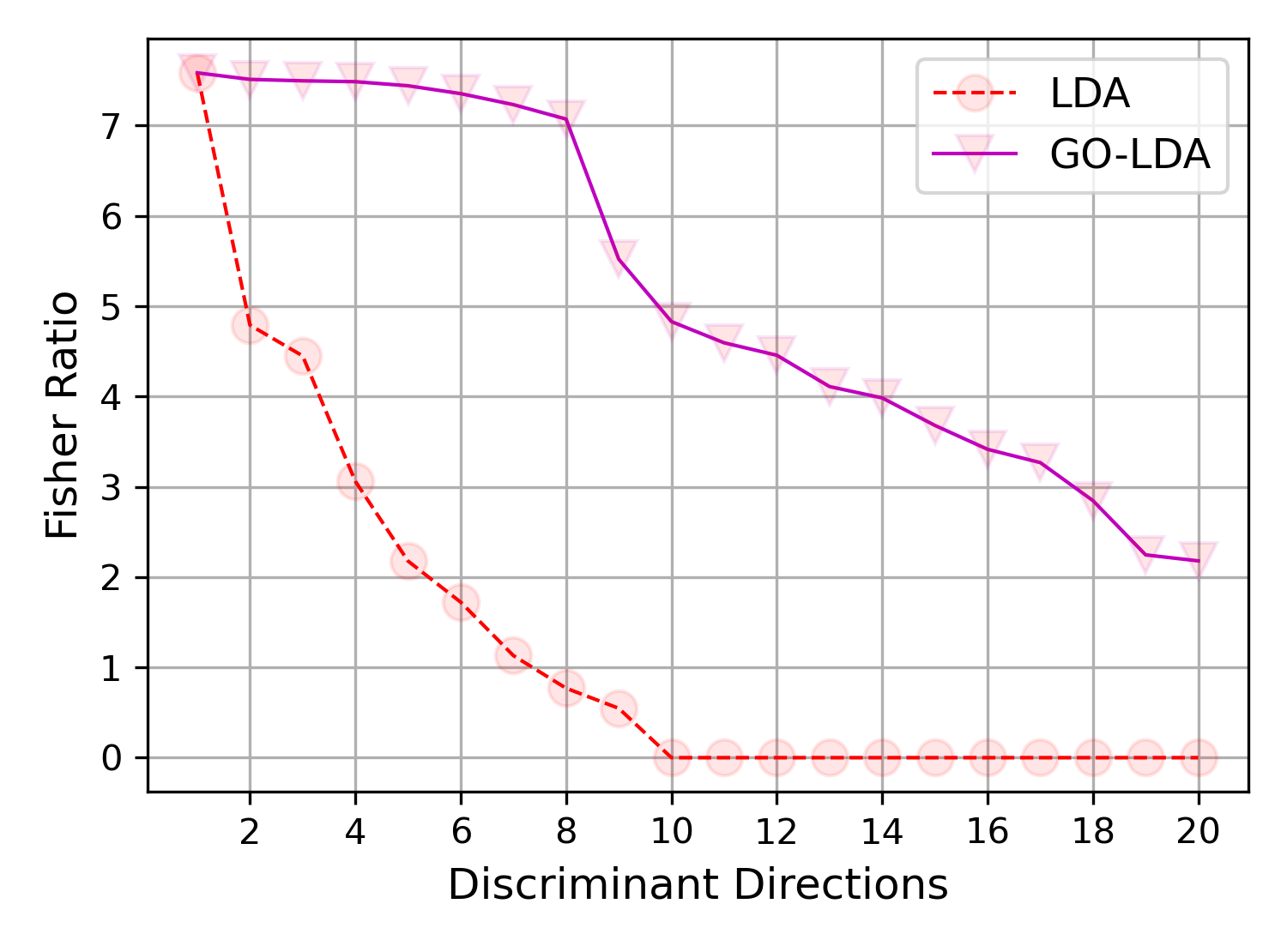}
\caption{Comparison between the classic-LDA and GO-LDA in terms of Fisher ratio corresponding to their individual discriminant directions on the {\tt Handwritten Digits} data set. The performance of the classic-LDA and GO-LDA is represented by the dashed and solid lines, respectively. It demonstrates that GO-LDA achieves a much higher level of discriminant ability compared to the classic-LDA even beyond the limit of the discriminant directions (i.e., eight directions) of the classic-LDA.}
\label{fisher_ratio}
\end{figure}

\subsection{Discriminant ability}
We finally illustrate the discriminant ability of the discriminant directions derived by our proposed GO-LDA and make comparison to the classic-LDA in terms of the Fisher ratio and projections on discriminant directions.

\subsubsection{Fisher ratio on discriminant directions}
The Fisher criterion is the objective function of LDA methods. Below we compare the Fisher ratio achieved by the discriminant directions of the classic-LDA and GO-LDA, see Fig. \ref{fisher_ratio}. The {\tt Handwritten Digits} classification data set \cite{Dua:2019} containing 10 classes in 64-dimensional feature space (i.e., pixel values of $8$$\times$$8$ images) is used in Fig. \ref{fisher_ratio}. Hence the classic-LDA only has nine discriminant directions available; in contrast, our GO-LDA is not restricted by the number of classes and can derive as many discriminant directions as the dimension of the feature space. Fig. \ref{fisher_ratio} clearly shows that the classic-LDA yields discriminant directions along which the Fisher ratio decreases rapidly; in other words, the discriminant ability of the classic-LDA is indeed quite limited. In contrast, our GO-LDA can derive more (optimal according to Theorem \ref{thm:go-lda-ui}) discriminant directions and retain much higher levels of discriminant ability even beyond the limit of the nine discriminant directions of the classic-LDA.

\subsubsection{Projections on discriminant directions}
To visually validate the optimal discriminant ability of GO-LDA, we investigate the projections on GO-LDA's discriminant directions and make comparison to the classic-LDA.
The {\tt Wine} data set, a benchmark classification problem containing 3 classes in 13-dimensional feature space taken from the {\tt UCI} ML repository \cite{Dua:2019}, is used here. Hence the classic-LDA will only be able to give two meaningful discriminant directions, whereas GO-LDA can derive 13 discriminant directions (i.e., the dimension of the feature space).

\begin{figure*}
  \centering
  \begin{subfigure}[b]{0.95\textwidth}
    \centering
    \includegraphics[width=\textwidth]{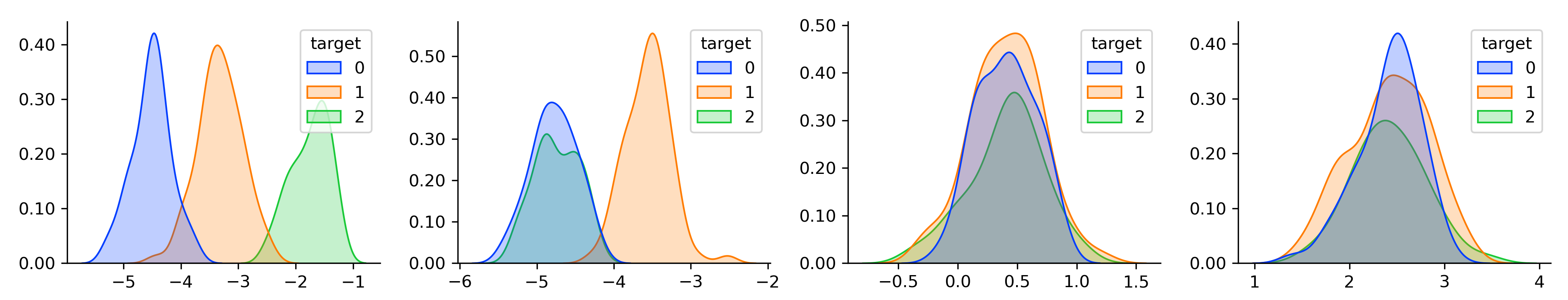}
    \caption{Classic-LDA}
    \label{fig:subfigure_a}
  \end{subfigure} \\[0.3cm]
  \begin{subfigure}[b]{0.95\textwidth}
    \centering
    \includegraphics[width=\textwidth]{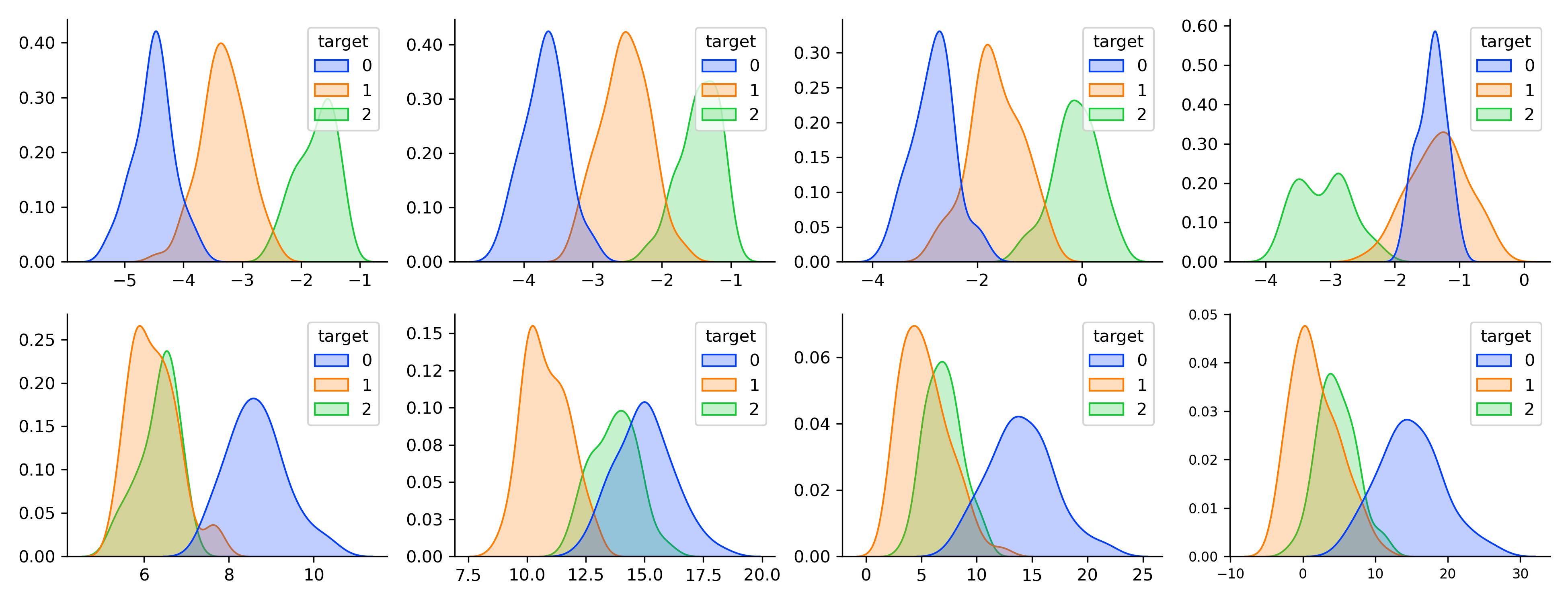}
    \caption{GO-LDA}
    \label{fig:subfigure_b}
  \end{subfigure}
\caption{Comparison between the classic-LDA and GO-LDA in terms of projections on their discriminant directions. The three-class {\tt Wine} data set with 13 features is used in this illustration. Panel (a) gives the projections of the {\tt Wine} data set on the classic-LDA's discriminant directions, i.e., $\boldsymbol{v}_{1}$ and $\boldsymbol{v}_{2}$, plus two more directions from solving the same generalised eigenvalue problem of the classic-LDA, respectively. Panel (b) gives the projections of the {\tt Wine} data set on the first eight GO-LDA's discriminant directions, i.e., $\boldsymbol{u}_{1}$ to $\boldsymbol{u}_{8}$, respectively.
The projections clearly show that GO-LDA delivers significantly better discriminant ability compared to the classic-LDA. In particular, the optimality of discrimination in GO-LDA is preserved for all the eight shown discriminant directions ({\it cf.} the classic-LDA can only find two discriminant directions for the three-class {\tt Wine} data set, with its second direction not optimal). A detailed description is given in the main text. A scatter plot in the two dimensional subspace consisting of $\boldsymbol{u}_{4}$ and $\boldsymbol{u}_{5}$ is shown in Appendix \ref{Appendix-d4andd5}. }
  \label{Wine}
\end{figure*}

Projections of the data onto the discriminant directions obtained by the classic-LDA and GO-LDA are shown in Fig. \ref{Wine}. From Fig. \ref{Wine} (a), we see that the projections of the three classes of the {\tt Wine} data set can be separated by the first discriminant direction (i.e., $\boldsymbol{v}_{1}$ or $\boldsymbol{u}_{1}$) of the classic-LDA and GO-LDA. However, the projections of two classes overlap along the classic-LDA's second discriminant direction  $\boldsymbol{v}_{2}$. Fig. \ref{Wine} (a) (see the right two plots) also shows the projections of the {\tt Wine} data set on two more directions obtained by solving the generalised eigenvalue problem of the classic-LDA, indicating that all of these directions are not useful and carry no discriminant information. In contrast, the GO-LDA's performance shown in Fig. \ref{Wine} (b) clearly demonstrates that all the GO-LDA's discriminant directions carry important discriminant information. In detail, the projections of the {\tt Wine} data set can be well separated by the first three GO-LDA's discriminant directions ({\it cf.} only the first discriminant direction of the classic-LDA can do so). Moreover, the following five GO-LDA's discriminant directions, i.e., $\boldsymbol{u}_{4}$ to $\boldsymbol{u}_{8}$, as shown in 
Fig. \ref{Wine} (b), can also achieve high separability after projecting the {\tt Wine} data set on them ({\it cf.} the directions derived by solving the generalised eigenvalue problem of the classic-LDA deliver no separability after $\boldsymbol{v}_{1}$ and $\boldsymbol{v}_{2}$). In particular, two classes of the {\tt Wine} data set slightly overlap after projecting onto the GO-LDA's discriminant directions $\boldsymbol{u}_{4}$ and $\boldsymbol{u}_{5}$ for example, but the overlap classes are different, indicating that each of the GO-LDA's discriminant directions carries different discriminant information, which benefits from the orthogonality between the GO-LDA's discriminant directions. Hence, interestingly, a combination of e.g. $\boldsymbol{u}_{4}$ and $\boldsymbol{u}_{5}$ in a two-dimensional projection can also separate all the three classes of the {\tt Wine} data set, see Appendix \ref{Appendix-d4andd5}.

The above illustrations showed the great performance of GO-LDA and the surprisingly limited performance of the classic-LDA in terms of discriminant ability. In the next section, we conduct a comprehensive set of experiments and comparisons to further validate GO-LDA's great discriminant ability and its importance in various applications.

\begin{table*}[h]
\caption{
Classification performance comparison between PCA, classic-LDA and GO-LDA using their individual principle/discriminant directions.
Six data sets all from the {\tt UCI} ML repository \cite{Dua:2019} are tested, with a quadratic classifier. Recall that $C$, $N$ and $M$ respectively represent the number of classes, the number of samples and the number of features of every sample in each data set. Up to $K=15$ principle/discriminant directions are used for PCA and GO-LDA. 
}
\label{table:individual}
\centering
\setlength{\tabcolsep}{0.8mm}
\begin{tabular}{c|c|c||c|c|c|c|c|c|c|c|c|c|c|c}
\toprule
\multirow{2}{*}{Data}&\multirow{2}{*}{$C$/$N$/$M$}&\multirow{2}{*}{Method}&\multicolumn{12}{c}{Accuracy on Different Principle/Discriminant Direction}
\\
&  & &1st&2nd&3rd&4th&5th&6th&7th&8th&9th&10th&$\cdots$&15th
\\
\midrule \midrule
\multirow{3}{*}{\tt IrisPlants} 
&\multirow{3}{*}{3/150/4} &PCA&0.90&\textbf{0.93}&0.40&0.27
\\
&&LDA&\textbf{1.0}&0.50&N/A&N/A
\\
&&GO-LDA&\textbf{1.0}&0.8&\textbf{0.90}&\textbf{0.80}
\\
\midrule
\multirow{3}{*}{\tt ThyroidGland}
&\multirow{3}{*}{3/215/5} &PCA&0.79&\textbf{0.97}&0.82&0.70&0.72
\\
&&LDA&\textbf{0.95}&0.79&N/A&N/A&N/A
\\
&&GO-LDA&\textbf{0.95}&0.88&\textbf{0.86}&\textbf{0.74}&\textbf{0.86}
\\
\midrule
\multirow{3}{*}{\tt Glass}
&\multirow{3}{*}{6/214/9}&PCA&0.46&0.51&0.42&0.46&0.37&0.39&0.37&0.35&0.36
\\
&&LDA&\textbf{0.65}&0.39&0.51&0.30&0.42&N/A&N/A&N/A&N/A
\\
&&GO-LDA&\textbf{0.65}&\textbf{0.69}&\textbf{0.69}&\textbf{0.58}&\textbf{0.51}&\textbf{0.49}&\textbf{0.47}&\textbf{0.40}&\textbf{0.40}
\\
\midrule
\multirow{3}{*}{\tt Wine}
&\multirow{3}{*}{3/178/13}&PCA&0.64&0.55&0.39&0.80&0.47&0.39&0.36&0.36&0.39&0.42
\\
&&LDA&\textbf{0.89}&0.69&N/A&N/A&N/A&N/A&N/A&N/A&N/A&N/A
\\
&&GO-LDA&\textbf{0.89}&\textbf{0.86}&\textbf{0.88}&\textbf{0.81}&\textbf{0.72}&\textbf{0.67}&\textbf{0.67}&\textbf{0.69}&\textbf{0.64}&\textbf{0.67}
\\
\midrule
\multirow{3}{*}{\tt Landsat}
&\multirow{3}{*}{6/6435/36} &PCA&0.47&0.64&0.57&0.22&0.33&0.25&0.29&0.24&0.26&0.23&$\cdots$&0.26
\\&&LDA&\textbf{0.55}&0.66&0.47&0.38&0.22&N/A&N/A&N/A&N/A&N/A&$\cdots$&N/A
\\
&&GO-LDA&\textbf{0.55}&\textbf{0.73}&\textbf{0.64}&\textbf{0.62}&\textbf{0.63}&\textbf{0.62}&\textbf{0.53}&\textbf{0.59}&\textbf{0.52}&\textbf{0.45}&$\cdots$&\textbf{0.46}
\\
\midrule
\multirow{3}{*}{\tt Handwritten Digits}
&\multirow{3}{*}{10/1797/64}  &PCA&0.17&0.40&0.35&0.34&0.26&0.32&0.24&0.22&0.25&0.22&$\cdots$&0.15
\\
&&LDA&\textbf{0.46}&0.41&0.34&0.29&0.26&0.28&0.26&0.22&0.20&N/A&$\cdots$&N/A
\\
&&GO-LDA&\textbf{0.46}&\textbf{0.46}&\textbf{0.47}&\textbf{0.48}&\textbf{0.45}&\textbf{0.46}&\textbf{0.46}&\textbf{0.36}&\textbf{0.39}&\textbf{0.42}&$\cdots$&\textbf{0.32}
\\
\midrule
\end{tabular}
\end{table*}

\section{Experimental Results}
\label{sec:experiments}
To showcase the effectiveness and importance of GO-LDA, we illustrate classification problems using both individual and multiple discriminant directions by carrying out extensive empirical work on a total of twenty benchmark data sets spanning different data types, numbers of classes and class imbalances. Throughout we compare the performance of our GO-LDA against the classic-LDA. Further, for completeness, we also compare with PCA by projecting the data onto the same number of principal components as that of the discriminant directions. Classifiers are built on those subspaces formed by principle/discriminant directions. 

\textbf{Data}. Among the twenty benchmark data sets, eleven are taken from the widely used {\tt UCI} ML repository \cite{Dua:2019}, i.e., {\tt IrisPlants}, {\tt BreastTissue}, {\tt ForestTypeMapping}, {\tt Glass}, {\tt Handwritten Digits}, {\tt Landsat}, {\tt Nursery}, {\tt ThyroidGland}, {\tt UrbanLandCover},
{\tt Vowel} and {\tt Wine}; five are taken from the {\tt KEEL} repository \cite{derrac2015keel} with high imbalance across classes, i.e., {\tt contraceptive},
{\tt Ecoli}, {\tt Hayes-Roth}, {\tt New-Thyroid} and {\tt Yeast}; two are face recognition data sets, i.e., {\tt LFW} \cite{LFWTech} and {\tt ORL} \cite{ORL}; 
and the rest two are medical data sets, i.e., {\tt BrainTumor} \cite{althnian2021impact} and  {\tt DeepDrid} \cite{DeepDRi}. Some necessary characteristics (e.g. the values of $C$, $N$ and $M$) of the data sets are given along with the corresponding results in Tables \ref{table:individual}--\ref{table:imbalance}. 

\textbf{Setting}. 
To make the comparisons more convincing, we choose different classifiers (i.e., $k$-nearest neighbour, linear and quadratic) to act on the subspaces.
To illustrate our work on inference from medical images, we first transfer the image data through a pre-trained deep neural network (i.e., ResNet18) into a fixed vector representation. The medical problems we consider here are characterised by the data scarcity scenario and thus only training deep image analysis models is generally insufficient. The projection techniques with discriminant abilities like PCA, classic-LDA and GO-LDA are essential.
In the transferred space, images are represented in a 512-dimensional feature space. The two medical problems are related to brain imaging (i.e., the {\tt BrainTumor} data set) \cite{althnian2021impact} and diabetic retinopathy (i.e., the {\tt DeepDrid} data set) \cite{DeepDRi}. In the former, 592 images are used for training and 148 for test; and with the latter, 529 images are used for training and 133 for test. Their figures are chosen to give a training set slightly higher than the image embedding dimensions (i.e., 512).

\subsection{Results on individual discriminant directions}
Table \ref{table:individual} gives the comparison between PCA, classic-LDA and GO-LDA on six data sets from the {\tt UCI} ML repository \cite{Dua:2019}, using quadratic classifiers on projections of the data onto different principle/discriminant directions taken one at a time. Note again that the number of the discriminant directions of the classic-LDA is limited by the number of classes $C$, which is significantly less than the number of the principle/discriminant directions of PCA and GO-LDA (which can go as large as the number of the features $M$).

The results in Table \ref{table:individual} show that GO-LDA outperforms the classic-LDA and PCA by a large margin. In detail, for the first principle/discriminant direction, both GO-LDA and classic-LDA achieve much higher classification accuracy than PCA, indicating the dramatic discriminant ability of GO-LDA and classic-LDA against PCA (which is variance preserving of the whole data but lack of discriminant ability). After that, all the GO-LDA's discriminant directions can achieve significantly better results than that of the classic-LDA, showing GO-LDA's discriminant optimality. In particular, after the third direction, all the GO-LDA's discriminant directions also achieve significantly better results than the principle directions of PCA. Another great advantage of GO-LDA shown from the results is that all the discriminant directions carry important discriminant information, i.e., their discriminant ability slightly decreases (which is what we expect according to the theory in Theorem \ref{thm:go-lda-ui}) when going to higher directions but in a quite slow manner ({\it cf.} the classic-LDA's discriminant ability drops sharply to not applicable after the $(C-1)$ direction).

\subsection{Results on discriminant subspaces}
Let $\Omega_l$ denote the discriminant subspace formed by the first $l$ principle/discriminant directions of PCA, classic-LDA or GO-LDA. Therefore, for PCA and GO-LDA, $l$ is the dimensionality of $\Omega_l$ and can be as large as $M$, while for the classic-LDA, $l$ can only be as large as $(C-1)$, which is generally much smaller than $M$. All the quantitative results with uncertainties in the tables below are estimated by ten-fold cross-validation.

Table \ref{table:UCIdata} gives the comparison between PCA, classic-LDA and GO-LDA  using the $k$-nearest neighbour classifier (with $k$ set to 1 for simplicity) on projections of the data onto discriminant subspaces $\Omega_l$ in terms of the mean classification accuracy (MCA).
The table shows results on five different {\tt UCI} problems, two face recognition and two medical problems. Consistent results are obtained in Table \ref{table:UCIdata}. For the results on the discriminant subspace $\Omega_{C-1}$, apart from a single outlier (i.e., results on the {\tt ForestTypeMapping} data set with discriminant subspace $\Omega_3$), the performance of the classic-LDA and GO-LDA outperforms PCA dramatically, and GO-LDA outperforms the classic-LDA. For the results on the discriminant subspace $\Omega_l$ where $l > C-1$, i.e., the subspace formed by using more principle/discriminant directions than the limit of the classic-LDA, GO-LDA achieves better results than PCA for all the cases including the case of the {\tt ForestTypeMapping} data set, again demonstrating GO-LDA's optimality of discriminant ability. In particular, we find that when more discriminant directions are used, GO-LDA's discriminant ability increases in six of the nine problems, sometimes substantially e.g. see the case of the {\tt DeepDrid} problem. This indicates that the data may have useful information over and beyond the first $(C-1)$ directions that GO-LDA can discover but the classic-LDA is constrained to find.

\begin{table*}[t]
\caption{
Classification performance comparison between PCA, classic-LDA and GO-LDA using their discriminant subspaces ($\Omega_l$) in terms of the MCA (mean classification accuracy).
Nine data sets (containing five different {\tt UCI} problems, two face recognition and two medical problems) are tested, with the $k$-nearest neighbour classifier where $k$ is set to 1 for simplicity. Uncertainties are estimated by ten-fold cross-validation. The results on the {\tt ForestTypeMapping} data set have no uncertainty since its training and test data sets are fixed by default. Recall that $\Omega_l$ represents the discriminant subspace formed by the first $l$ principle/discriminant directions of PCA, classic-LDA and GO-LDA individually. 
}
\label{table:UCIdata}
\centering
\setlength{\tabcolsep}{0.8mm}
\begin{tabular}{c|c|c||p{1.3cm}p{1.1cm} |p{1.3cm} p{1.1cm}}
\toprule
\multirow{2}{*}{Data} & \multirow{2}{*}{$C$/$N$/$M$} & \multirow{2}{*}{Method} & \multicolumn{4}{c}{ MCA on Discriminant Subspaces}
\\
 &  & & \multicolumn{2}{c|}{$\Omega_{C-1}$} & \multicolumn{2}{c}{$\Omega_{l}$} 
\\
\midrule \midrule
&&PCA&0.94$\pm$0.05 & (on $\Omega_{2}$)  & 0.96$\pm$0.05 & (on $\Omega_{4}$)
\\
{\tt IrisPlants }
& 3/150/4 &LDA&0.96$\pm$0.05 & (on $\Omega_{2}$) &N/A &
\\
&&GO-LDA&\textbf{0.98$\pm$0.03} & (on $\Omega_{2}$) &\textbf{0.96$\pm$0.04} &   (on $\Omega_{4}$)
\\
\midrule
&&PCA&0.21$\pm$0.12 & (on $\Omega_{8}$) &0.42$\pm$0.10 &  (on $\Omega_{20}$)
\\
{\tt UrbanLandCover }
&9/168/147&LDA&0.31$\pm$0.11 & (on $\Omega_{8}$) &N/A &
\\
&&GO-LDA&\textbf{0.43$\pm$0.12}& (on $\Omega_{8}$) &\textbf{0.56$\pm$0.08}  &  (on $\Omega_{20}$)
\\
\midrule
&&PCA&0.52$\pm$0.19 & (on $\Omega_{5}$) &0.51$\pm$0.10 &   (on $\Omega_{9}$)
\\
{\tt BreastTissue }
&6/106/9&LDA&0.55$\pm$0.14 & (on $\Omega_{5}$) &N/A &
\\
&&GO-LDA&\textbf{0.61$\pm$0.16} & (on $\Omega_{5}$) &\textbf{0.52$\pm$0.13} &   (on $\Omega_{9}$)
\\
\midrule
&&PCA&\textbf{0.82} & (on $\Omega_{3}$) &0.83 &   (on $\Omega_{10}$)
\\
{\tt ForestTypeMapping }
&4/523/27&LDA&0.79 & (on $\Omega_{3}$) &N/A &
\\
&&GO-LDA&0.76 & (on $\Omega_{3}$) &\textbf{0.84} & (on $\Omega_{10}$)
\\
\midrule
&&PCA&0.51$\pm$0.08 & (on $\Omega_{3}$) &0.89$\pm$0.03 &  (on $\Omega_{8}$)
\\
{\tt Nursery }
&4/12960/8&LDA&0.86$\pm$0.02 & (on $\Omega_{3}$) &N/A &
\\
&&GO-LDA&\textbf{0.90$\pm$0.01} & (on $\Omega_{3}$) &\textbf{0.95$\pm$0.02} &  (on $\Omega_{8}$)
\\
\midrule
\midrule
&&PCA&0.30$\pm$0.03 & (on $\Omega_{4}$) &0.46$\pm$0.04 &   (on $\Omega_{10}$)
\\
{\tt LFW }
&5/1140/1850&LDA&\textbf{0.66$\pm$0.02}& (on $\Omega_{4}$) &N/A &
\\
&&GO-LDA&\textbf{0.66$\pm$0.02}& (on $\Omega_{4}$) &\textbf{0.74$\pm$0.04} &   (on $\Omega_{10}$)
\\
\midrule
&&PCA&0.78$\pm$0.02 & (on $\Omega_{39}$) & 0.97$\pm$0.02 &(on $\Omega_{50}$)
\\
{\tt ORL }
&40/400/10304&LDA&0.98$\pm$0.01 & (on $\Omega_{39}$) &N/A
\\
&&GO-LDA&\textbf{0.99$\pm$0.02} & (on $\Omega_{39}$) &\textbf{0.99$\pm$0.01} &  (on $\Omega_{50}$)
\\
\midrule
\midrule
&&PCA&0.38$\pm$0.01 & (on $\Omega_{3}$) &0.39$\pm$0.01 &  (on $\Omega_{10}$)
\\
{\tt BrainTumor }
&4/800/512&LDA&0.53$\pm$0.02 & (on $\Omega_{3}$) &N/A &
\\
&&GO-LDA&\textbf{0.57$\pm$0.01} & (on $\Omega_{3}$) &\textbf{0.59$\pm$0.02} & (on $\Omega_{10}$)
\\
\midrule
&&PCA&0.27$\pm$0.01 & (on $\Omega_{4}$) &0.44$\pm$0.05 & (on $\Omega_{20}$)
\\
{\tt DeepDrid }
&5/662/512&LDA&0.27$\pm$0.04& (on $\Omega_{4}$) &N/A &
\\
&&GO-LDA&\textbf{0.29$\pm$0.04} & (on $\Omega_{4}$) &\textbf{0.50$\pm$0.05} &  (on $\Omega_{20}$)
\\
\midrule
\end{tabular}
\end{table*}

\begin{table*}[t]
\caption{
Classification performance comparison between PCA, classic-LDA and GO-LDA using their discriminant subspaces ($\Omega_l$) in terms of the MCA.
Three data sets all from the {\tt UCI} ML repository \cite{Dua:2019} are tested, with a linear classifier. Please refer to the work reported in \cite{duin2004linear} for comparison against published results. 
}
\label{table:compare3}
\centering
\setlength{\tabcolsep}{0.8mm}
\begin{tabular}{c|c|c||c p{1.0cm}|c p{1.1cm}|p{1.25cm} p{1.1cm}}
\toprule
\multirow{2}{*}{Data} & \multirow{2}{*}{$C$/$N$/$M$} & \multirow{2}{*}{Method} & \multicolumn{6}{c}{ MCA on Discriminant Subspaces}
\\
 &  & & \multicolumn{2}{c|}{ $\Omega_{l}, l\le 3$}  & \multicolumn{2}{c|}{$\Omega_{C-1}$} & \multicolumn{2}{c}{$\Omega_{l}, l = \min\{M, 10\}$}
\\
\midrule \midrule
&&PCA&0.45$\pm$0.09 & (on $\Omega_{3}$) &0.54$\pm$0.10 & (on $\Omega_{5}$) &0.61$\pm$0.07 & (on $\Omega_{9}$)
\\
{\tt Glass }
& 6/214/9 &LDA&0.42$\pm$0.11 & (on $\Omega_{3}$)  &0.51$\pm$0.09 & (on $\Omega_{5}$) &N/A
\\
&&GO-LDA&\textbf{0.53$\pm$0.09} & (on $\Omega_{3}$) &\textbf{0.57$\pm$0.06} & (on $\Omega_{5}$) &\textbf{0.63$\pm$0.07} & (on $\Omega_{9}$)
\\
\midrule
&&PCA&0.55$\pm$0.09 & (on $\Omega_{3}$) &0.49$\pm$0.10 & (on $\Omega_{5}$) &0.49$\pm$0.09 & (on $\Omega_{10}$)
\\
{\tt Landsat }
& 6/6435/36 &LDA&0.71$\pm$0.07 & (on $\Omega_{3}$) &0.69$\pm$0.07 & (on $\Omega_{5}$) & N/A
\\
&&GO-LDA&\textbf{0.75$\pm$0.07} & (on $\Omega_{3}$) &\textbf{0.77$\pm$0.05} & (on $\Omega_{5}$) &\textbf{0.74$\pm$0.06} & (on $\Omega_{10}$)
\\
\midrule
&&PCA&0.41$\pm$0.06 & (on $\Omega_{2}$)  &0.51$\pm$0.03 & (on $\Omega_{10}$) &0.51$\pm$0.03 & (on $\Omega_{10}$)
\\
{\tt Vowel }
& 11/990/10 &LDA&0.49$\pm$0.05 & (on $\Omega_{2}$)  &0.53$\pm$0.05 & (on $\Omega_{10}$) &0.53$\pm$0.05 & (on $\Omega_{10}$)
\\
&&GO-LDA&\textbf{0.50$\pm$0.04} & (on $\Omega_{2}$) &\textbf{0.54$\pm$0.05} & (on $\Omega_{10}$) &\textbf{0.54$\pm$0.05} & (on $\Omega_{10}$)
\\
\midrule
\end{tabular}
\end{table*}

Table \ref{table:compare3} shows the results obtained similar to the way in Table \ref{table:UCIdata} but with a linear classifier on different discriminant subspaces and comparing with previously published results by Duin et al. \cite{duin2004linear}. Three data sets from the {\tt UCI} ML repository are tested. Similar and consistent results are obtained. In particular, our GO-LDA achieves the best results compared to PCA and classic-LDA for all the cases by a large margin.
We also note that going from a discriminant subspace $\Omega_l$ formed by a small number of discriminant directions, e.g., $l \le 3$, to one formed by a large number of discriminant directions, e.g., $l = C-1$, GO-LDA can already achieve higher accuracies than that reported in \cite{duin2004linear}. 


Finally, we make similar comparisons on the highly imbalanced data sets from the {\tt KEEL} repository \cite{derrac2015keel}. The results are given in Table \ref{table:imbalance} using the $k$-nearest neighbour classifier. Again, our GO-LDA outperforms the classic-LDA for all the cases and outperforms PCA for most of the cases (i.e., eight out of the ten cases). Interestingly, compared to the previous results, the performance difference between PCA, classic-LDA and GO-LDA is not that significant on the highly imbalanced data sets, and the performance gains are minor when considering subspaces formed by more principle/discriminant directions.
This might be because the dominated classes could be predicted well by just using the first few number of principle/discriminant directions, and the classes being dominated are insignificant for the MCA due to their small size. Further investigation of highly imbalanced data is of great interest for future work.

\begin{table*}[t]
\caption{
Classification performance comparison between PCA, classic-LDA and GO-LDA using their discriminant subspaces ($\Omega_l$) in terms of the MCA.
Five highly imbalanced data sets (indicated by the imbalanced ratio in the third column of the table below) from the {\tt KEEL} repository \cite{derrac2015keel} are tested, with the $k$-nearest neighbour classifier. }
\label{table:imbalance}
\centering
\setlength{\tabcolsep}{0.8mm}
\begin{tabular}{c|c|c|c||c p{1.0cm}|p{1.25cm} p{1.0cm}}
\toprule
\multirow{2}{*}{Data} & \multirow{2}{*}{$C$/$N$/$M$} & Imbalanced & \multirow{2}{*}{Method} & \multicolumn{4}{c}{ MCA on Discriminant Subspaces}
\\
 &  & Ratio & & \multicolumn{2}{c|}{$\Omega_{C-1}$} & \multicolumn{2}{c}{$\Omega_{M}$}
\\
\midrule \midrule
&&&PCA&\textbf{0.46$\pm$0.04} & (on $\Omega_{2}$) &0.45$\pm$0.03 & (on $\Omega_{9}$) 
\\
{\tt contraceptive }
& 3/1473/9 &1.89&LDA&0.42$\pm$0.05 & (on $\Omega_{2}$)  &N/A &
\\
&&&GO-LDA&0.42$\pm$0.04 & (on $\Omega_{2}$)  &\textbf{0.45$\pm$0.02} & (on $\Omega_{9}$)
\\
\midrule
&&&PCA&0.69$\pm$0.11 & (on $\Omega_{2}$)  &\textbf{0.75$\pm$0.09} & (on $\Omega_{4}$)
\\
{\tt Hayes-Roth }
& 3/132/4 &1.7&LDA&0.70$\pm$0.13 & (on $\Omega_{2}$)  &N/A &
\\
&&&GO-LDA&\textbf{0.70$\pm$0.09} & (on $\Omega_{2}$)  &0.73$\pm$0.09 & (on $\Omega_{4}$)
\\
\midrule
&&&PCA&0.91$\pm$0.05 & (on $\Omega_{2}$)  &0.90$\pm$0.06 & (on $\Omega_{5}$)
\\
{\tt New-Thyroid }
& 3/215/5 &4.84&LDA&0.95$\pm$0.04 & (on $\Omega_{2}$) &N/A &
\\
&&&GO-LDA&\textbf{0.95$\pm$0.03} & (on $\Omega_{2}$)  &\textbf{0.96$\pm$0.03}  & (on $\Omega_{5}$)
\\
\midrule
&&&PCA&0.81$\pm$0.05 & (on $\Omega_{7}$) &0.81$\pm$0.05  & (on $\Omega_{7}$)
\\
{\tt Ecoli }
& 8/336/7 &71.5&LDA&0.79$\pm$0.05 & (on $\Omega_{7}$) &N/A &
\\
&&&GO-LDA&\textbf{0.83$\pm$0.05} & (on $\Omega_{7}$) &\textbf{0.83$\pm$0.05}  & (on $\Omega_{7}$)
\\
\midrule
&&&PCA&0.52$\pm$0.04 & (on $\Omega_{9}$) &0.51$\pm$0.04  & (on $\Omega_{8}$)
\\
{\tt Yeast }
& 10/1484/8 &23.15&LDA&\textbf{0.52$\pm$0.02} & (on $\Omega_{9}$) &N/A &
\\
&&&GO-LDA&
\textbf{0.52$\pm$0.02} & (on $\Omega_{9}$) &\textbf{0.52$\pm$0.02}  & (on $\Omega_{8}$)
\\
\midrule
\end{tabular}
\end{table*}

\section{Conclusion}
\label{sec:conclusion}
We in this paper proposed GO-LDA, 
deriving how discriminant directions can be sequentially extracted for multiclass data analysis and pattern classification problems through maximising Fisher criterion and retaining orthogonality to previously computed ones. The textbook solution to computing such a discriminant subspace has been by solving a generalized eigenvalue problem, i.e., the classic-LDA. Surprisingly, this solution, which has been in the literature for several decades, does not preserve the mutual orthogonality of resulting directions; nor do the resulting individual directions hold high discrimination. Moreover, the classic solution is restricted to finding a subspace, the dimensionality of which is limited by the number of classes in the problem. Our derivation of GO-LDA in this paper sequentially optimizes a set of discriminant directions that, while preserving discrimination and mutual orthogonality, can naturally go beyond the limit imposed by the rank of the between-class scatter matrix. The excellent performance of GO-LDA was supported by illustrative examples showing the objective function (i.e., Fisher ratio) and distributions of projections as well as an extensive set of multiclass classification experiments taken from machine learning benchmark data sets with thorough comparisons. 

Problems we consider for discriminant analysis may be seen as relatively small by standards of the very large-scale problems such as computer vision arising in modern machine learning. However, they are quite pertinent to validate the property and power of the proposed GO-LDA in both theoretical and practical manners. It is also well established that small data-size problems are still of paramount interest in applications such as medical diagnostics, where either due to the prevalence of a complex disease or due to restrictions arising from privacy issues that limit the amount of data available for training and validating models. Even with settings that demand non-linear classification boundaries, it is possible to have fixed nonlinear transformations using for example pre-trained deep neural networks followed by linear discrimination models acting on their feature spaces. The two medical problems we used for illustration in this paper fall precisely in this space, emphasising the importance of the study we report. In the present work, we continue in this area of medical inference, incorporating uncertainty via probabilistic modelling in the derivation of discriminant subspaces.  

Ubiquitous applications of GO-LDA are evident. For future avenues, it is of great interest to investigate highly imbalanced data and transfer the essence of GO-LDA to other LDA variants.


%

\appendices


\section{Derivation of
$\boldsymbol{d}_{2}$ for the Binary Problem}\label{Appendix-dd}
This Appendix derives an important step in the derivation of the second discriminant direction $\boldsymbol{d}_{2}$ for the binary problem, which is not shown in \cite{foley1975optimal,sammon1970optimal}.

Maximising the objective function in Eq. \eqref{Eqn:d2} in Section \ref{sec:methodology} with respect to $\boldsymbol{d}_{2}$ can be addressed by solving
\begin{equation}\label{a.1}
\frac{2 \tilde{\boldsymbol{S}}_{\rm B} \boldsymbol{d}_{2}}{\boldsymbol{d}_{2}^{\top} {\boldsymbol{S}}_{\rm W} \boldsymbol{d}_{2}}-\frac{2 \boldsymbol{d}_{2}^{\top} \tilde{\boldsymbol{S}}_{\rm B} \boldsymbol{d}_{2} {\boldsymbol{S}}_{\rm W} \boldsymbol{d}_{2}}{(\boldsymbol{d}_{2}^{\top} {\boldsymbol{S}}_{\rm W} \boldsymbol{d}_{2})^{2}}-\lambda \boldsymbol{d}_{1}=0 .
\end{equation}
Note that $\tilde{\boldsymbol{S}}_{\rm B} = \boldsymbol{s}_{\rm b} \boldsymbol{s}_{\rm b}^{\top}$. Substituting it into Eq. \eqref{a.1} yields
\begin{equation}
\frac{2 \boldsymbol{s}_{\rm b} \boldsymbol{s}_{\rm b}^{\top} \boldsymbol{d}_{2}}{\boldsymbol{d}_{2}^{\top} {\boldsymbol{S}}_{\rm W} \boldsymbol{d}_{2}}-\frac{2 \boldsymbol{d}_{2}^{\top} \boldsymbol{s}_{\rm b} \boldsymbol{s}_{\rm b}^{\top} \boldsymbol{d}_{2} {\boldsymbol{S}}_{\rm W} \boldsymbol{d}_{2}}{(\boldsymbol{d}_{2}^{\top} {\boldsymbol{S}}_{\rm W} \boldsymbol{d}_{2})^{2}}-\lambda \boldsymbol{d}_{1}=0 .
\label{a.2}
\end{equation}
Let $\kappa = \boldsymbol{s}_{\rm b}^{\top} \boldsymbol{d}_{2} / \boldsymbol{d}_{2}^{\top} {\boldsymbol{S}}_{\rm W} {\boldsymbol{d}_{2}}$, which is a scalar. Eq. \eqref{a.2} can be rewritten as

\begin{equation}
2 \kappa \boldsymbol{s}_{\rm b} - 2 \kappa^{2} {\boldsymbol{S}}_{\rm W} \boldsymbol{d}_{2}-\lambda \boldsymbol{d}_{1}=0.
\label{a.3}
\end{equation}
Then we have
\begin{equation}
\boldsymbol{d}_{2}=\frac{1}{\kappa} {\boldsymbol{S}}_{\rm W}^{-1}\left(\boldsymbol{s}_{\rm b}-\frac{\lambda}{2 \kappa} \boldsymbol{d}_{1}\right).
\label{A.4}
\end{equation}
Since $\boldsymbol{d}_{1} = {\boldsymbol{S}}_{\rm W}^{-1} \boldsymbol{s}_{\rm b}$, we get
\begin{equation}
\boldsymbol{d}_{2} =\frac{1}{\kappa}\left({\boldsymbol{S}}_{\rm W}^{-1}-\frac{\lambda}{2 \kappa}\left({\boldsymbol{S}}_{\rm W}^{-1}\right)^{2}\right) \boldsymbol{s}_{\rm b} .
\label{eqn:d2-temp}
\end{equation}

Let $S_{11}=\boldsymbol{d}_{1}^{\top} {\boldsymbol{S}}_{\rm W}^{-1} \boldsymbol{d}_{1}$. Since $\boldsymbol{d}_{1}^{\top} \boldsymbol{d}_{2}=0$, we have 
\begin{equation}
\boldsymbol{d}_{1}^{\top} \boldsymbol{d}_{2}=\frac{1}{\kappa} \boldsymbol{d}_{1}^{\top} {\boldsymbol{S}}_{\rm W}^{-1} \boldsymbol{s}_{\rm b}-\frac{\lambda}{2 \kappa^{2}} \boldsymbol{d}_{1}^{\top} {\boldsymbol{S}}_{\rm W}^{-1} \boldsymbol{d}_{1} = 0,
\label{a.6}
\end{equation}
which gives
\begin{equation}
\frac{1}{\kappa} \boldsymbol{d}_{1}^{\top}\boldsymbol{d}_{1}-\frac{\lambda}{2 \kappa^{2}} S_{11} = 0 .
\end{equation}
Therefore, 
\begin{equation}
\frac{\lambda}{2 \kappa}=\frac{\boldsymbol{d}_{1}^\top \boldsymbol{d}_{1}}{S_{11}}.
\end{equation}
Since
\begin{equation}
\begin{aligned}
S_{11} &= \boldsymbol{d}_{1}^{\top} {\boldsymbol{S}}_{\rm W}^{-1} \boldsymbol{d}_{1} \\
& = \boldsymbol{s}_{\rm b}^{\top} {\boldsymbol{S}}_{\rm W}^{-1} {\boldsymbol{S}}_{\rm W}^{-1} {\boldsymbol{S}}_{\rm W}^{-1} \boldsymbol{s}_{\rm b} ,\\
& = \boldsymbol{s}_{\rm b}^{\top}\left({\boldsymbol{S}}_{\rm W}^{-1}\right)^{3} \boldsymbol{s}_{\rm b},
\end{aligned}
\end{equation}
and $\boldsymbol{d}_{1}^{\top}\boldsymbol{d}_{1} = \boldsymbol{s}_{\rm b}^{\top}\left({\boldsymbol{S}}_{\rm W}^{-1}\right)^{2} \boldsymbol{s}_{\rm b}$, 
we have
\begin{equation}
\frac{\lambda}{2 \kappa}=\frac{\boldsymbol{s}_{\rm b}^{\top}\left({\boldsymbol{S}}_{\rm W}^{-1}\right)^{2} \boldsymbol{s}_{\rm b}}{\boldsymbol{s}_{\rm b}^{\top}\left({\boldsymbol{S}}_{\rm W}^{-1}\right)^{3} \boldsymbol{s}_{\rm b}}.
\end{equation}
Substituting it into Eq. \eqref{eqn:d2-temp}, we have
\begin{equation}
\boldsymbol{d}_{2} =\frac{1}{\kappa}\left({\boldsymbol{S}}_{\rm W}^{-1} - \frac{\boldsymbol{s}_{\rm b}^{\top}({{\boldsymbol{S}}_{\rm W}}^{-1})^{2} \boldsymbol{\boldsymbol{s}_{\rm b}}}{\boldsymbol{s}_{\rm b}^{\top}({{\boldsymbol{S}}_{\rm W}}^{-1})^{3} \boldsymbol{s}_{\rm b}} ({{\boldsymbol{S}}_{\rm W}}^{-1})^{2}\right) {\boldsymbol{s}_{\rm b}} .
\end{equation}
Normalising $\boldsymbol{d}_{2}$ above, we then complete the derivation.

\section{Combination of GO-LDA's Discriminant Directions}\label{Appendix-d4andd5}
\begin{figure}[h]
\centering
\includegraphics[width=3.0in]{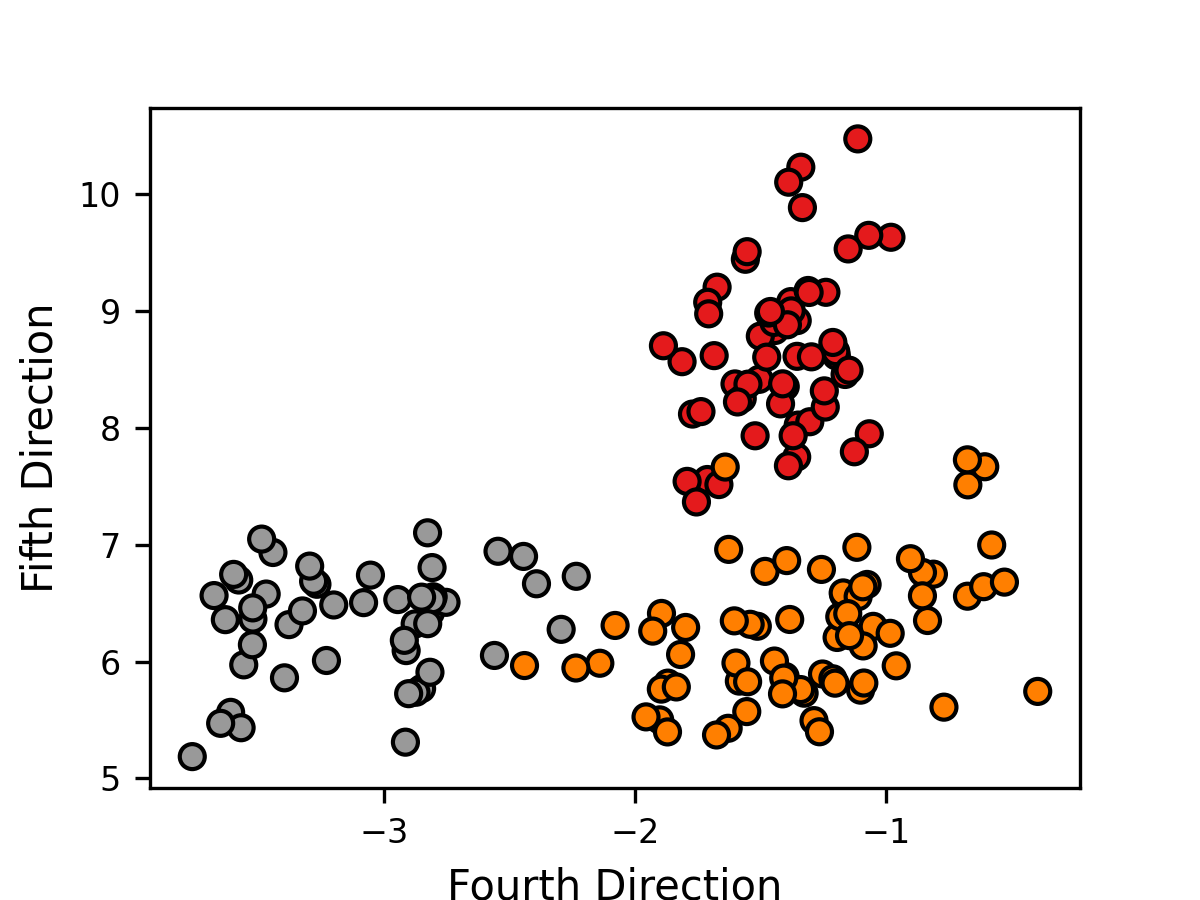}
\caption{
Projection of the three-class {\tt Wine} data set on two of GO-LDA's discriminant directions. It shows that, when projected along the two directions $\boldsymbol{u}_{4}$ and $\boldsymbol{u}_{5}$, this three-class data set continues to show separation between classes. This would not have been possible with the classic-LDA's solution from which only two discriminant directions could have been extracted; therefore, individuals seeking to acquire additional discriminant directions for further investigation would find the fourth and fifth directions of limited applicability in the classic-LDA.}
\label{d4andd5}
\end{figure}

When extracting multiple mutually orthogonal discriminant directions with GO-LDA, an intriguing observation is that even when individual directions do not necessarily separate all the classes, their combinations do as illustrated in Fig. \ref{d4andd5}. The data set used here is from the {\tt Wine} problem, a benchmark classification problem sourced from the {\tt UCI} ML repository \cite{Dua:2019}, consisting of 178 13-dimensional samples that are categorised into three distinct classes. 

\ifCLASSOPTIONcompsoc
  \section*{Acknowledgments}
\else
  \section*{Acknowledgment}
\fi

MN's contribution to the work was partially funded by Engineering and Physical Sciences Research Council (EPSRC) grant ``Early detection of contact distress for enhanced performance monitoring and predictive inspection of machines'' (EP/S005463/1).

\ifCLASSOPTIONcaptionsoff
  \newpage
\fi



%

\bibliographystyle{IEEEtran}
\bibliography{bibliography}

\end{document}